\documentclass{arXiv}

\usepackage{type1cm}        
\usepackage{makeidx}         
\usepackage{graphicx}        
\usepackage{multicol}        
\usepackage[bottom]{footmisc}

\usepackage{newtxtext}       %
\usepackage{newtxmath}       
\usepackage{bm}

\usepackage[first=-100, last=100]{lcg}
\newcommand{\random}{\rand\arabic{rand}}

\usepackage{amsmath}
\DeclareMathOperator*{\argmax}{arg\,max}

\usepackage{algorithm}
\usepackage[noend]{algpseudocode}
\algnewcommand\algorithmicforeach{\textbf{for each}}
\algdef{S}[FOR]{ForEach}[1]{\algorithmicforeach\  #1\ \algorithmicdo}

\usepackage{caption}
\makeatletter
\def\IfClass#1#2#3{\@ifundefined{opt@#1.cls}{#3}{#2}}
\makeatother

\IfClass{svmult} {
    \usepackage{xcolor}
    
}{}

\makeindex             

\begin{document}

\IfClass{svmult} {
    \title*{Quality-Diversity Optimization: a novel branch of stochastic optimization}
    \author{Konstantinos Chatzilygeroudis \and Antoine Cully \and Vassilis Vassiliades \and Jean-Baptiste Mouret}
    \authorrunning{Chatzilygeroudis \and Cully \and Vassiliades \and and Mouret}
    \institute{Konstantinos Chatzilygeroudis \at Computer Technology Institute \& Press ``Diophantus'' (CTI), GR-26504, University Campus, Patras, Greece, \email{kchatzil@cti.gr}
    \and Antoine Cully \at Adaptive \& Intelligent Robotics Lab, Imperial College London, South Kensington Campus London SW7 2AZ, UK, \email{a.cully@imperial.ac.uk}
    \and Vassilis Vassiliades \at Research Centre on Interactive Media, Smart Systems and Emerging Technologies, Dimarcheio Lefkosias, Plateia Eleftherias, Nicosia 1500, Cyprus, \email{v.vassiliades@rise.org.cy}
    \and Jean-Baptiste Mouret \at Inria, CNRS, Universit\'e de Lorraine, LORIA, F-54000 Nancy, France, \email{jean-baptiste.mouret@inria.fr}}
    %
    %
}{
    \title{Quality-Diversity Optimization:\\a novel branch of stochastic optimization}
	\titlesize{19}
	\author{Konstantinos Chatzilygeroudis$^\dag$,
        Antoine Cully$^\ddag$,
		Vassilis Vassiliades$^*$,
		and Jean-Baptiste Mouret$^\diamond$}
	\authorshort{Chatzilygeroudis, Cully, Vassiliades, and Mouret}
	\affiliations{$^\dag$Computer Technology Institute \& Press ``Diophantus'' (CTI), GR-26504, University Campus, Patras, Greece\\
        $^\ddag$Adaptive \& Intelligent Robotics Lab, Imperial College London, South Kensington Campus London SW7 2AZ, UK\\
        $^*$Research Centre on Interactive Media, Smart Systems and Emerging Technologies, Dimarcheio Lefkosias, Nicosia 1500, Cyprus\\
		$^\diamond$Inria, CNRS, Universit\'e de Lorraine, LORIA, F-54000 Nancy, France}
}

\maketitle



\begin{abstract}
    Traditional optimization algorithms search for a single global optimum that maximizes (or minimizes) the objective function. Multimodal optimization algorithms search for the highest peaks in the search space that can be more than one. Quality-Diversity algorithms are a recent addition to the evolutionary computation toolbox that do not only search for a single set of local optima, but instead try to illuminate the search space. In effect, they provide a holistic view of how high-performing solutions are distributed throughout a search space. The main differences with multimodal optimization algorithms are that (1) Quality-Diversity typically works in the behavioral space (or feature space), and not in the genotypic (or parameter) space, and (2) Quality-Diversity attempts to fill the whole behavior space, even if the niche is not a peak in the fitness landscape. In this chapter, we provide a gentle introduction to Quality-Diversity optimization, discuss the main representative algorithms, and the main current topics under consideration in the community. Throughout the chapter, we also discuss several successful applications of Quality-Diversity algorithms, including deep learning, robotics, and reinforcement learning.
\end{abstract}

\section{Introduction}
\label{sec:intro}
Optimization has countless uses in engineering, from designing mechanical parts~\cite{sigmund200199} to controling robots~\cite{mayne2000constrained,escande2014hierarchical}. In an ideal situation, the user of an optimization algorithm would write the cost function that corresponds to the problem at hand, select the right optimization algorithm, and get \emph{the} perfect solution. Unfortunately, many real-world optimization problems are not easily captured by a simple cost function, which is why it is often required to add new terms to the cost functions and/or tune the parameters until the optimization algorithm finally reaches an acceptable solution. Consequently, optimization in engineering is often an iterative process that requires many runs of the optimizers and many changes of the cost function.

Morover, even after fine-tuning the cost function, the optimized solution is rarely used as the final, optimal solution. In practice, optimization is most often used at the beginning of the design process to explore various options and examine the trade-offs that are inherent to the domain~\cite{Bradner2014}. This calls for algorithms that are designed as exploration tools more than as pure optimization tools.

Quality-Diversity (QD) optimization algorithms address this challenge: instead of searching for the optimum of the cost function, they provide a large set of high-performing solutions (typically a few thousands) that differ according to a few user-defined features of interest. The user can then pick the high-performing solutions that they deem as the most interesting according to their own knowledge, like aesthetics or easiness of manufacturing. He/she can also use the resulting set to better understand how features of solutions influence performance. For instance, one of the features might have no influence on the cost function, or another one might need to be below a given threshold to attain acceptable performance.

For instance, QD algorithms have been used to find gait parameters for a 6-legged robot~\cite{cully2013behavioral,chatzilygeroudis2018reset,duarte2018evolution}. In that task, instead of finding parameters to go forward, a QD algorithm can find \emph{in a single run} parameters so that the robot can reach any point in its vicinity (e.g., walking forward, backward, left, etc.). For each direction, the algorithm will find a different set of high-performing parameters; however, close directions are likely to have similar solutions, which means that it may be more efficient to optimize for all the directions simulatenously. In particular, if a set of parameters is tested and makes the robot turn right, it will be useless to go forward, but will be a promising solution to turn right. In addition, these parameters may be good ``stepping stones'' to go forward (or backward, or left, etc.), that is, a useful intermediate step in the optimization process.

Numerous QD algorithms have been proposed during recent years, which are mostly based on the principles of evolutionary algorithms. This chapter gives an overview of these algorithms and introduce several recent ideas to increase the data-efficiency, scale to high-dimensional spaces. Application examples are discussed all along the chapter.
\section{Problem Formulation}
\label{sec:problem_formulation}
 
We assume that the objective function returns both the fitness value $f_{\boldsymbol{\theta}}$ and a behavioral descriptor (or a feature vector) $\boldsymbol{b}_{\boldsymbol{\theta}}$~\cite{mouret2020quality}:
\begin{align}
    \label{eq:qd_f}
    f_{\boldsymbol{\theta}}, \boldsymbol{b}_{\boldsymbol{\theta}} \leftarrow f(\boldsymbol{\theta})
\end{align}
The behavioral desciptor (BD) typically describes \emph{how} the solution solves the problem, while the fitness value $f_{\boldsymbol{\theta}}$ quantifies \emph{how well} it solves it. For example, the BD can be the curvature of a 3D design, its volume, or the trajectory of a robot, while the fitness values would be the aerodynamic drag, the energy consumption, or the distance to the target state.

Without loss of generality, we assume hereafter that the fitness function is maximimized. Let us define by $\mathcal{B}$ the feature space, the goal in QD optimization is to find for each point $\boldsymbol{b}\in\mathcal{B}$ the parameters, $\boldsymbol{\theta}$, with the maximum fitness value:
\begin{align}
    \label{eq:qd_opt}
    \forall\boldsymbol{b}\in\mathcal{B}\quad&\boldsymbol{\theta}^* = \argmax_{\boldsymbol{\theta}}f_{\boldsymbol{\theta}}\nonumber\\
    &\text{s.t.}\quad\boldsymbol{b} = \boldsymbol{b}_{\boldsymbol{\theta}}
\end{align}

For instance, if the feature descriptors are discretized, the outcome of a QD algorithm can be a table. See Table~\ref{tab:feature} for an example:
\begin{table}[H]
    \centering
    \begin{tabular}{|c c | c | c c c c|}
        \hline
         \multicolumn{2}{|c|}{BD} & fitness & \multicolumn{4}{c|}{Parameter values}\\
        \hline
        \random&\random& \random & \random & \random & \random & \random \\
        \random&\random& \random & \random & \random & \random & \random\\
        \multicolumn{2}{|c|}{ \vdots} &  \vdots & \multicolumn{4}{c|}{ \vdots}\\
        \random&\random& \random & \random & \random & \random & \random \\
        \hline
    \end{tabular}
    \caption{\label{tab:feature}An example of discretized feature descriptor.}
\end{table}
When the BD is 2-dimensional, this result is usually displayed as a colored image or heatmap.
    
At first sight, QD algorithms look like multitask optimization \cite{mouret2020quality}, that is, solving an optimization problem for each combination of features. However, first $\mathcal{B}$ can be continuous, which would mean an infinite number of problems; second, we do not know the BD before calling the fitness function, which explains why the QD problem can be viewed as a set of optimizations \emph{constrained} by each BD.

The central hypothesis of QD algorithms is that solving this set of problems is likely to be faster if they are all solved together than by independent constrained optimizations. Intuitively, it is indeed likely that high-performing solutions for close feature descriptors will be close, therefore sharing information between the optimizations can be beneficial. In addition, independent constrained optimization would be especially wasteful in a black-box optimization context because a candidate solution that has not the right features would be discarded, whereas it could be useful for a different feature combination.
\subsection{Collections of Solutions}
\label{sec:collections}
The outcome of QD optimization is a set of solutions. This set of solutions, also called ``collection'', ``archive'', or ``map'', is expanded, improved, and refined during the optimization process. Each point in this collection represents a different ``solution type'' or ``species''.
In practice, two solutions with similar behavioral descriptor will be considered to be of the same solution type and will compete together to be maintained in the collection. Necessarily, the notion of similarity is defined by using a hyper-parameter that sets the tolerance used to determine when two descriptors are different or similar. This hyper-parameter defines a sort of ``resolution'' in the behavioral descriptor space: only one solution will occupy a certain region of the space.

The simplest way to implement this segmentation of the BD space is by discretizing it into a grid, in which each cell of the grid corresponds to one type of solution (i.e., to one BD location). This approach is used by the MAP-Elites algorithm \cite{mouret2015illuminating}, one of the most used QD algorithm. In this case, the collection is a grid (or multidimensional array) and the goal of the algorithm is to fill every cell of that grid with the best possible solution. However, it is possible to avoid this discretization and replace it with distance thresholds or local density estimates (section ~\ref{sec:optim_collection}).


\subsection{How do we measure the performance of a QD algorithm?}
\label{sec:metrics}

The overall performance of a QD algorithm is defined by the quality of the produced collection of solutions according to two criteria:
\begin{enumerate}
    \item the performance of the solution found for each type of solutions (how much we have optimized);
    \item the coverage of the behavior space (how much of the feature space is covered).
\end{enumerate}
%

The first criterion (performance) is obvious to compute: depending on the application or use-case, we can compute the mean, median or the sum of the individual fitness values in the collection. The second criteria (coverage) can be more challenging to evaluate when the behavior/feature space is not discretized. If we are in low-dimensional spaces, we can discretize the behavior space arbitrarily and compute the percentage of the bins filled by the algorithms~\cite{pugh2016quality}. Alternatively, if we are operating in a high-dimensional space, we can resort to density metrics, like the average distance of the k-nearest neighbors. A third option is to define a distance threshold between behavioral descriptions, and then compute a similar filling percentage as in the low-dimensional space case.
\section{Optimizing a collection of solutions}
\label{sec:optim_collection}

In this section, we begin with the introduction of \emph{Multidimensional Archive of Phenotypic Elites} (MAP-Elites)~\cite{mouret2015illuminating}, and then present a modern categorization of QD algorithms~\cite{cully2018quality} that makes it to define many variants of QD algorithms, including more advanced algorithms.
\subsection{MAP-Elites}
\label{sec:map_elites}
MAP-Elites takes inspiration from evolutionary algorithms: at each iteration, MAP-Elites alters copies of solutions that are already in the grid to form new solutions (see Algo.~\ref{algo:map_elites}). The alterations are done with mutation and cross-over operators like in traditional evolutionary algorithms. The new solutions are evaluated and then potentially added to the cell corresponding to their BD. If the cell is empty, the solution is added to the grid. Otherwise, only the best solution is kept in the cell.
%
\begin{algorithm}
    \caption{MAP-Elites algorithm}\label{algo:map_elites}
    \begin{algorithmic}[1]
        \Procedure{MAP-Elites}{$[n_1,...,n_d]$}
            \State $\mathcal{A} \longleftarrow$ create\_empty\_archive($[n_1,...,n_d]$)
            \For {$i=1\to G$} \Comment{\emph{Initialization: $G$ random $\boldsymbol{\theta}$}}
                \State $\boldsymbol{\theta} = $ random\_solution()
                \State \textsc{add\_to\_archive}($\boldsymbol{\theta}, \mathcal{A}$)
            \EndFor
            \For {$i=1\to I$} \Comment{\emph{Main loop, $I$ iterations}}
                \State $\boldsymbol{\theta} = $ selection($\mathcal{A}$)
                \State $\boldsymbol{\theta}' = $ variation($\boldsymbol{\theta}$)
                \State \textsc{add\_to\_archive}($\boldsymbol{\theta}', \mathcal{A}$)
            \EndFor
            \State \textbf{return} $\mathcal{A}$
        \EndProcedure
        \Procedure{add\_to\_archive}{$\boldsymbol{\theta}, \mathcal{A}$}
            \State $(p,\mathbf{b}) \longleftarrow $ evaluate($\boldsymbol{\theta}$)
            \State $c \longleftarrow $ get\_cell\_index($\mathbf{b}$)
            \If {$\mathcal{A}(c) = null$ or $\mathcal{A}(c).p < p$}
                \State $\mathcal{A}(c) \longleftarrow p, \boldsymbol{\theta}$
            \EndIf
        \EndProcedure
    \end{algorithmic}
\end{algorithm}

MAP-Elites has been successfully employed in many domains. For instance, it has been used to produce: behavioral repertoires that enable robots to adapt to damage in a matter of minutes~\cite{cully2015robots,tarapore2016different}, perform complex tasks~\cite{duarte2018evolution}, or even adapt to damage while completing their tasks~\cite{chatzilygeroudis2018reset}; morphological designs for walking ``soft robots'', as well as behaviors for a robotic arm~\cite{mouret2015illuminating}; neural networks that drive simulated robots through mazes~\cite{pugh2015confronting}; images that ``fool'' deep neural networks~\cite{nguyen2015deep}; ``innovation engines'' able to generate images that resemble natural objects~\cite{nguyen2015innovation}; and 3-D-printable objects by leveraging feedback from neural networks trained on 2-D images~\cite{lehman2016creative}.

The main strength of MAP-Elites is its simplicity to understand and to implement. However, depending on the domain, it is sometimes difficult to discretize the feature/behavior space (for instance, in high dimensions or when the bounds are highly irregular). It is also possible to bias the selection in different way: the uniform selection among the elites of the original gives suprisingly good results but several ideas have been investigated to select parents in other ways that can make the convergence faster.

In the following section, we present a modern categorization of QD algorithms that make it possible to describe most QD algorithms within the same algorithmic framework.
\subsection{A Unified Framework}
\label{sec:qd_framework}
Cully and Demiris~\cite{cully2018quality} proposed a unified formulation of QD algorithms that see all QD algorithms as an instantiation of a single high-level algorithm (Algo.~\ref{algo:QD}). In this formulation, the main axes of variation of all QD algorithms are: (1) the type of container, that is, how the data are gathered and ordered into a collection, (2) the type of the selection operator, that is, how the solutions are selected to be altered in the next generation, and (3) the type of scores that are being computed in order for the container and the selection operator to work.

In particular, after a random initialization, the execution of a QD-algorithm based on this framework follows four steps that are repeated:
\begin{itemize}
    \item the selection operator produces a new set of individuals that will be altered in order to form the new batch of evaluations,
    \item the individuals are evaluated and their performance and BD are recorded,
    \item each of these individuals is then potentially added to the container, according to the solutions already in the collection,
    \item finally, several scores, like the novelty, the local competition, or the curiosity score, are updated.
\end{itemize}

These four steps repeat until a stopping criterion is reached (typically, a maximum number of iterations) and the algorithm outputs the collection stored in the container.
In the following sections, we will detail different variants of the containers, the selection operators, and of the most widely used scores.
%
%
\begin{algorithm*}
    \small
    \caption{QD-Optimization algorithm ($I$ iterations)}
    \label{algo:QD}
    \begin{algorithmic}
    \State $\mathcal{A} \leftarrow \emptyset$\Comment{\emph{Creation of an empty container.}}
    \For{iter $  = 1\to I$} \Comment{\emph{The main loop repeats during $I$ iterations.}}
    \If{iter $== 1$} \Comment{\emph{Initialization.}}
    \State $\mathcal{P}_{\text{parents}}\leftarrow $ random() \Comment{\emph{The first $2$ batches of individuals are generated randomly.}}
    \State $\mathcal{P}_{\text{offspring}}\leftarrow $ random()
    \Else \Comment{\emph{The next controllers are generated using the container and/or the previous batch.}}
    \State $\mathcal{P}_{\text{parents}}\leftarrow $ selection($\mathcal{A}$, $\mathcal{P}_{\text{offspring}}$) \Comment{\emph{Selection of a batch of individuals from the container and/or the previous batch.}}
    \State $\mathcal{P}_{\text{offspring}}\leftarrow $ variation($\mathcal{P}_{\text{parents}}$) \Comment{\emph{Creation of a randomly modified copy of $\mathcal{P}_{\text{parents}}$ (mutation and/or crossover).}}
    \EndIf
    \ForEach {$\boldsymbol{\theta}$ $\in \mathcal{P}_{\text{offspring}}$}
    \State $\{f_{\boldsymbol{\theta}},\boldsymbol{b}_{\boldsymbol{\theta}}\}\leftarrow$ $f(\boldsymbol{\theta})$ \Comment{\emph{Evaluation of the individual and recording of its descriptor and performance.}}
    \If{\textsc{add\_to\_container}($\boldsymbol{\theta}$, $\mathcal{A}$)} \Comment{\emph{``\textsc{add\_to\_container}'' returns true if the individual has been added to the container.}}
    \State \textsc{update\_scores}(parent($\boldsymbol{\theta}$), Reward, $\mathcal{A}$) \Comment{\emph{The parent might get a reward.}}
    \Else
    \State \textsc{update\_scores}(parent($\boldsymbol{\theta}$), -Penalty, $\mathcal{A}$) \Comment{\emph{Otherwise, it might get a penalty.}}
    \EndIf
    \EndFor
    \textsc{update\_container}($\mathcal{A}$) \Comment{\emph{Update of the attributes of all the individuals in the container (e.g. novelty score).}}
    \EndFor
    \State \Return $\mathcal{A}$
    \end{algorithmic}
\end{algorithm*}
\subsubsection{Containers}
\label{sec:containers}
The main purpose of a container is to gather all the solutions found so far into an ordered collection, in which only the best and most diverse solutions are kept.

One of the most popular container types in the literature is the \textit{$N-$dimensional grid structure}, which is the one that MAP-Elites is using. In this container, we simply discretize the behavior space in a grid, where each cell of the grid corresponds to one type of solution. Originally, the MAP-Elites grid was built with only one solution per cell. Of course, one can imagine having more individuals per cell (e.g.,~\cite{pugh2016quality} uses two individuals) in order to perform more complicated computations (e.g., for multi-objective optimization or noisy optimization~\cite{justesen2019map, flageat2020fast}). In high-dimensional behavior spaces, it is possible to use a Centroidal Voronoi Tessellation to define cells of identical volume regardless of the dimension (see Sec.~\ref{sec:high_dim_qd}).

An alternative container type is the \textit{distanced-based archive}. In this type of containers, the solutions are kept in an unstructured array by using their behavior descriptor and the Euclidean distance. In essence, the user specifies a threshold parameter and a new individual is added to the archive (a) if it is ``far away'' from all other solutions in the archive (its Euclidean distance greater than the user-defined threshold), or (b) if it is better than its closest(s) neighbor(s). In contrast with the grid container presented previously, the descriptor space here is not discretized and the structure of the collection autonomously emerges from the encountered solutions. However, in practice this container type requires a slightly more sophisticated maintenant mechanism to avoid the ``erosiion effect'' that may progressively remove solutions that are far from the rest of the collection in favor of solutions that are slightly closer but with a higher value, slowing down the overall optimization process.  
%
%
\subsubsection{Selection Operators}
\label{sec:selection_ops}
One of the keys for the success of QD algorithms, but also evolutionary algorithms in general, is the selection operators. The selection operators aim at answering the following question: given the current collection (or population), how do we sample or generate new individuals to be evaluated?
%

The most naive way of generating solutions is by \textit{randomly sampling the parameter space}, that is, by not using the current collection. This is unlikely to be effective as it makes the QD algorithms identical to random search.
%
%

The original MAP-Elites implementation \textit{randomly samples solutions from the ones already in the container}. This is a very simple strategy to implement and very computational efficient. However, one of its main drawbacks is that the selection pressure decreases as the number of solutions in the collection increases (the chance for a solution to be selected being inversely proportional to the number of solutions in the collection), which is likely to be ineffective with large collections.
%

Another interesting way of selecting new individuals is adopting \textit{a score-based weighting for the random sampling}. In this way, we can insert more weight on more ``interesting'' individuals based on a specific score, and bias the selection pressure towards desired behaviors or more ``interesting'' individuals. In QD optimization, we aim at devising scores that will improve the collection, that is, to either discover new behaviors or optimize the already discovered ones (see Sec.~\ref{sec:scores}).
%
%

So far, all the defined selection operators are operating on the stored container. In this case, the population of the QD algorithm is the container itself that \emph{evolves} and improves over time. One can imagine having \textit{multiple populations that evolve in parallel}. For example, Novelty Search with Local Competition (NSLC)~\cite{lehman2011evolving} uses two distinct populations, one as a container storing information about novelty (they call it \emph{novelty archive}) and one more traditional population storing information about performance and using it to perform the selection operation. 
\subsubsection{Population Scores}
\label{sec:scores}
Traditionally, evolutionary algorithms consider only the \textit{fitness value} (quality) to take decisions. Most of these approaches will struggle to identify diverse solutions, and might even fail at finding the global optimum. On the contrary, QD algorithms consider additional quantities in an attempt to better explore the behavior space (diversity).
%
%
%

One of the most widely used scores is the \textit{novelty score}. The novelty score attempts to put higher values to solutions that are more different than other solutions, and thus forcing the algorithm to keep a diverse set of solutions instead of many similar ones. The most common formulation for the novelty score is the average distance of the k-neareset neighbors in the behavior space. This technique is introduced by the novelty search algorithms~\cite{lehman2011abandoning,lehman2011evolving}.
%

Another idea is to reward individuals that produce offsprings that are novel enough or better than the individuals already in the container. In this way, the algorithm will prefer to sample individuals that will most likely produce new cells in the container or replace already occupied cells. The \textit{curiosity score} attempts to do this by modelling the probability of an individual to generate offsprings that will be added to the container. One practical implementation of the curiosity score (see Algo.~\ref{algo:QD}) is to begin with zero curiosity score for all individuals, and then each time that one of their offsprings gets added to the container, their curiosity score increases, whereas it decreases each time their offsprings do not enter the container.

Finally, Go-Explore~\cite{ecoffet2019go,ecoffet2020first} introduced the concept of attempting to \textit{expand the frontier of the search} by promoting the selection of newly discovered individuals. The intuition is that a newly discovered individual will most likely contain an interesting behavior that can potentially lead to novel regions of the search space. The authors practically implement this idea by introducing a score that is based on visit and selection counters.
%
%
\subsection{Considerations of Quality-Diversity Optimization}
\label{sec:example_qd}
%
Results from ~\cite{cully2018quality} indicate that the usage of curiosity score (see Sec.~\ref{sec:scores}) in a QD algorithm seems to be an effective way of keep exploring interesting regions of the behavior space while also locally optimizing the solutions. The performance of QD algorithms with the curiosity score both in distanced-based and grid-based containers was consistently better than many other variants~\cite{cully2018quality}. In particular, this type of algorithms were able to learn behavior repertoires for a six-legged robot to walk in any direction or walk in a straight line in many different ways. These results showcase that relying on individuals with a high-propensity to generate individuals that are added to the collection is a promising selection heuristic.

Another quite interesting takeaway is the fact that using the uniform random selection operator (over the container), MAP-Elites' selection operator, is very effective both when used in distance-based and grid-based containers (the original MAP-Elites implementation)~\cite{cully2018quality,cully2015robots,mouret2015illuminating}. This showcases the strength of selecting candidates to reproduce from the elites (contained in the archive), rather than randomly generating new ones.

Additionally, Novelty-Search with Local Competition~\cite{lehman2011evolving} is less effective than curiosity-based QD instances and QD instances with the MAP-Elites' selection operator~\cite{cully2018quality}. This finding advocates that evolving the whole collection of solutions is more effective than splitting it in multiple populations with different characteristics and functionalities.

Finally, the distance-based containers produce containers with smaller numbers of individuals. This of course depends on the choice of the threshold for the distance comparisons, but also highlights the need for better structure of the containers as it is often not very easy to tune these parameters. In Sec.~\ref{sec:high_dim_qd}, we discuss different types of containers to handle some of these limitations.
\section{Origins and Related Work}
\subsection{Searching for Diverse Behaviors}
\label{sec:related}
Quality-Diversity algorithms mainly originate from the desire to evolve multiple and diverse behaviors in the evolutionary robotics community~\cite{lehman2011evolving,lehman2011abandoning,mouret2012encouraging}. In particular, they build on top of Novelty Search (NS)~\cite{lehman2011abandoning}, and Novelty-Search with Local Competition (NSLC)~\cite{lehman2011evolving} algorithms, which propose to search for novel solutions instead of high-performing ones. This proved to be particularly interesting to escape from deceptive regions in the search space (local optima) and reach eventually high-performing solutions~\cite{mouret2012encouraging}. Searching for novel solutions is done by rewarding solutions that are different from the previously encountered ones thanks to the \textit{novelty score} (described in section~\ref{sec:scores}). This score is implemented in practice by maintaining an archive, called the novelty archive, of the previously encountered solutions, which is then used to compute the average distance in the BD space between the contained solutions and any newly generated one. The archive is mainly used as a tool to compute the novelty score, while the actual outcome of the algorithms is their final population (like most evolutionary algorithms). It is also important to note that the archive was designed to quantify the coverage of the BD space (and thus compute the novelty score), not to produce a collection of high-performing and diverse solution.

Cully and Mouret proposed in the Behavioural-Repertoire Evolution algorithm (BR-Evolution)~\cite{cully2013behavioral} to consider the novelty archive of NSLC as the outcome of the algorithm by introducing mechanisms to only keep the best-performing solutions by replacing them when a better one is found. This idea of building a large collection of solutions that produce different behaviors while maximizing the performance of each type of behaviors is one of the very first instances of Quality-Diversity algorithms as known today. At the same time, Mouret et al. designed a simple algorithm to plot a figure showing the distribution of high-performing solutions over a given feature space (illuminating the fitness landscape)~\cite{clune2013evolutionary}. Surprisingly, this simple algorithm, named later MAP-Elites~\cite{mouret2015illuminating} was in practice very effective in evolving behavioural repertoires like the BR-Evolution algorithm~\cite{cully2015robots}. Shortly after, the concept of generating a collection of diverse and high-performing solutions has been formalized and named Quality-Diversity~\cite{pugh2015confronting, pugh2016quality}.
\subsection{Connections to Multimodal Optimization}
\label{sec:connections}
\begin{figure*}
    \centering
    \includegraphics[width=1\textwidth]{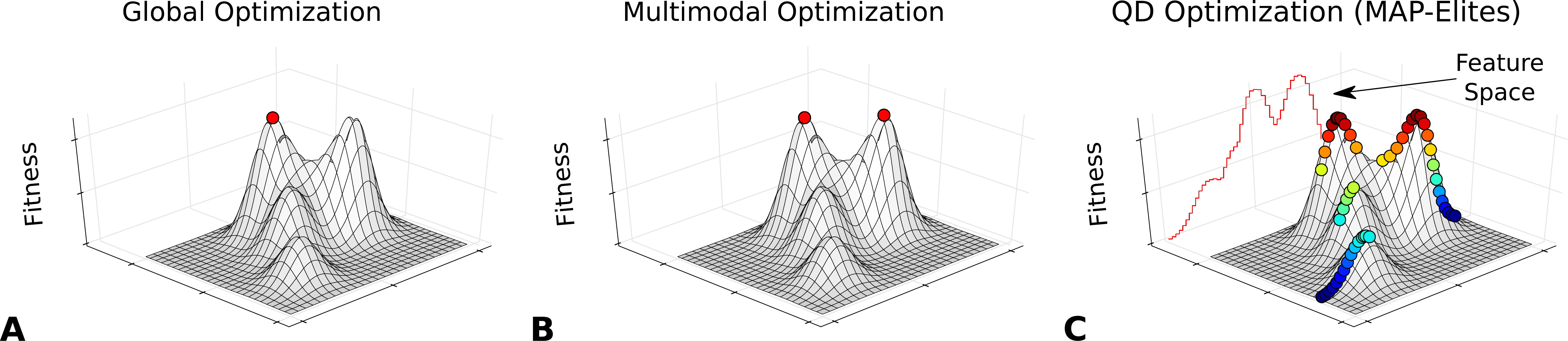}
    \caption{Difference between global, multimodal and QD optimization. The goal of global optimization algorithms is to find a single global optimum of the objective function (\textbf{A}). Multimodal optimization (MMO) algorithms aim to return multiple optima (\textbf{B}). QD optimization algorithms, such as MAP-Elites (\textbf{C}), discover significantly more solutions, each one being the highest-performing of a local neighborhood defined in some \textit{feature space} of interest. For finding the solutions of the function illustrated, we used (\textbf{A}) Covariance Matrix Adaptation Evolution Strategies~\cite{hansen2001completely}, (\textbf{B}) Restricted Tournament Selection~\cite{harik1995rts}, and (\textbf{C}) MAP-Elites~\cite{mouret2015illuminating}. Figure adapted from~\cite{vassiliades2018using}.}\label{fig_multimodal_QD}
\end{figure*}

%
Traditionally, the focus of optimization was to find a single, globally optimal solution of the objective function (Fig.~\ref{fig_multimodal_QD}A).
It is often the case, however, that (1) the objective function is highly nonlinear, which might cause even sophisticated gradient-free algorithms, such as evolution strategies, to converge to local optima~\cite{rudolph2001self}, and (2) the user would like from the optimization algorithm to return all local optima in order to choose between them (for example, this could be the case in engineering or design problems~\cite{lee1999niching, shir2007conceptual}).
Multimodal optimization (MMO) algorithms (e.g., see~\cite{goldberg1987genetic,mahfoud1995niching, harik1995rts, sareni1998fitness, das2011real, preuss2015multimodal,singh2006comparison,yin1993fast,petrowski1996clearing}) seek to address these issues, by employing various diversity maintenance techniques, also known as \emph{niching} methods, with the aim to return multiple solutions that correspond to the peaks of the search space (Fig.~\ref{fig_multimodal_QD}B).

Niching has a long history in evolutionary computation. One of the earliest attempts, was the \emph{preselection} method~\cite{cavicchio1970adaptive} in which an offspring replaces its least fit parent, if it has higher fitness. \emph{Crowding}~\cite{de1975analysis} was the first to propose the use of distances (to the best of our knowledge). \emph{Fitness sharing}~\cite{goldberg1987genetic} assumes that fitness is a precious resource that is shared among neighboring individuals, therefore, it reduces the fitness of individuals in densely populated regions. \emph{Clearing}~\cite{petrowski1996clearing} is a technique that has the same inspiration as sharing, but rather than lowering the fitness, it removes less fit individuals from the neighborhoods. In \emph{restricted tournament selection}~\cite{harik1995rts} an offspring competes with its closest individual in a randomly selected sample of the population. Clustering~\cite{yin1993fast} and multi-objective optimization~\cite{deb2010finding} were also proposed (among others) as a way to maintain diversity. In addition, niching techniques have been proposed for various nature-inspired optimization algorithms, such as particle swarm optimization~\cite{barrera2009review}. A comprehensive survey is outside the scope of this chapter, however, the interested reader can refer to~\cite{das2011real,preuss2015multimodal}.

In order to better highlight the similarities and differences between global and multimodal optimization (assuming maximization) let us formally define their objectives. Global optimization can be written as:
\begin{align}
    \label{eq:global_opt}
    \boldsymbol{\theta}^*_{g} = \argmax_{\boldsymbol{\theta} \in \Theta}f({\boldsymbol{\theta}})
\end{align}
or in set-builder notation:
\begin{align}
    \label{eq:global_opt_set}
    \left\{ \boldsymbol{\theta}^*_{g} \in \Theta : f(\boldsymbol{\theta}^*_{g}) > f(\boldsymbol{\theta}), \forall \boldsymbol{\theta} \in \Theta \right\} 
\end{align}
where $\Theta$ is the parameter space. In other words, global optimization algorithms aim to return a set that contains a single solution which is the globally optimal one. MMO can typically be expressed as:
\begin{align}
    \label{eq:multi_modal}
    \left\{ \boldsymbol{\theta}^*_{l} \in \Theta : f(\boldsymbol{\theta}^*_{l}) > f(\boldsymbol{\theta}), \forall \boldsymbol{\theta} \in \Theta, d(\boldsymbol{\theta},\boldsymbol{\theta}^*_{l})<\epsilon, \epsilon>0 \right\} 
\end{align}
where $d$ is a distance function; in other words, MMO aims to return the set of all locally optimal solutions, typically in parameter space.

Quality-Diversity optimization on the other hand aims at optimizing a different objective function that returns both a fitness value and a \emph{behavior descriptor} (see Sec.~\ref{sec:problem_formulation} and Eq.~\ref{eq:qd_f},~\ref{eq:qd_opt}). Alternatively, it can be expressed as:
\begin{align}
    \label{eq:qd_opt_set}
    \left\{ \boldsymbol{\theta}^*_{QD} \in \Theta : f(\boldsymbol{\theta}^*_{QD}) > f(\boldsymbol{\theta}), \forall \boldsymbol{\theta} \in \Theta, d(b(\boldsymbol{\theta}), b(\boldsymbol{\theta}^*_{QD}))<\epsilon, \epsilon>0 \right\} 
\end{align}
where $f : \Theta \rightarrow \mathbb{R}$, and $b : \Theta \rightarrow \mathcal{B}$. Observing Eq.~\ref{eq:qd_opt_set}, we can easily identify that the main difference of QD from multimodal optimization is in the focus on finding the optimal parameters for each point in the behavior space, as well as returning many more points than optima (Fig.~\ref{fig_multimodal_QD}C).
%
%

It is clear by now that QD algorithms attempt to solve a different problem than more traditional optimization techniques. It should also be obvious that it might be possible to use off-the-shelf (local or global) optimizers with restart procedures (or in parallel) to solve the multimodal optimization problem, however, not the QD problem. For example, if we instantiate parallel hill climbers from uniformly-spread random initial points, they might return the optima, however (1) some of them might return the same solution, and (2) the whole set of solutions will most probably \emph{not} be as diverse as the one returned by QD optimization algorithms. Two questions arise now: (1) can we use multimodal optimization algorithms to solve the QD problem? (2) can we use QD algorithms to solve the multimodal optimization problem?

Although there are not many works in the literature that investigate these questions, it has been shown that some multimodal optimization algorithms can perform as well as QD algorithms if set to compare distances in behavior space~\cite{vassiliades2017comparing}, while others fail at doing so. In particular, the clearing method~\cite{petrowski1996clearing} was able to solve QD problems, whereas Restricted Tournament Selection~\cite{harik1995rts} was not able to do so, mainly because of its strong focus on performance (rather than diversity) which makes it even lose certain local optima (Fig.~\ref{fig_multimodal_QD}B). This showcases the need for algorithms that attempt to solve the specific QD problem, and that the QD problem cannot be generically solved by other types of optimization methods. There are, however, a lot of intuitions to be taken from the multimodal optimization literature to improve QD algorithms.

Additionally, it is shown that QD algorithms can also work in high-dimensional parameter space~\cite{vassiliades2017comparison,vassiliades2018using}, and thus QD algorithms can be used to solve multimodal optimization problems. Typically, QD algorithms return many more solutions than the optima of the underlying search space, therefore, finding the optima in the returned set of solutions could potentially be done using some filtering technique (such as the nearest better clustering heuristic~\cite{preuss2005counteracting}).
\subsection{Connections to Multi-task Optimization}
As introduced in section \ref{sec:problem_formulation}, QD algorithms assume that the fitness function $f(\bm{\theta})$ returns both the fitness value $f_\theta$ and a behavioral descriptor $\bm{b}_\theta$:
\begin{equation}
    f_\theta, \bm{b}_\theta \leftarrow f(\bm{\theta})
\end{equation}
By contrast, multitask optimization considers a fitness function that is parameterized by a task descriptor $\bm{\tau}$ and returns the fitness value:
\begin{equation}
    f_{\theta,\tau} \leftarrow f(\bm{\theta}, \bm{\tau})
\end{equation}
The task descriptor might describe, for example, the morphology of a robot or the features of a game level; it is typically a vector of numbers that describes the parameters of the task.

The overall objective is to find, for each task $\bm{\tau}$, the solution $\bm{\theta}_\tau^*$ with the maximum fitness:
\begin{equation}
    \forall \bm{\tau} \in T, \bm{\theta}_\tau^* = \textrm{argmax}_{\bm{\theta}} \big(f(\bm{\theta}, \bm{\tau})\big)
\end{equation}

To our knowledge, only a few algorithms have been proposed for multitask optimization, mainly in the framework of Bayesian optimization \cite{pearce2018continuous}. However, the MAP-Elites algorithm was recently extended to solve multitask optimization problems \cite{mouret2020quality}. The general idea is to select the task $\bm{\tau}$ in the neighborhood of the parents, selected using the standard MAP-Elites procedure (uniform selection from the archive). The results show that Multi-task-MAP-Elites can outperform independent optimizations with the CMA-ES algorithm \cite{hansen2001completely}, especially in hard optimization problems, most probably because it can achieve a more global search by looking at all the tasks simultaneously \cite{mouret2020quality}.
\section{Current Topics}
\subsection{Expensive Objective Functions}
\label{sec:data_efficient_qd}
QD algorithms are designed for non-convex, black-box functions with many peaks. As such, they typically assume that the objective function can be queried millions of times. For instance, MAP-Elites is typically given a budget of 20 million evaluations to find about 15,000 effective gaits for a 6-legged robot, with 36 parameters that define the gait~\cite{cully2015robots,chatzilygeroudis2018reset}. Unfortunately, many interesting engineering problems, for example aerodynamics optimization, require simulating each candidate solution for minutes or even hours: in these engineering problems, calling the objective function millions of times is not possible, even when parallelizing on large clusters and multicore computers.

This challenge is common for all black-box optimization algorithms, if not for all the optimization algorithms. In such cases, the traditionnal approach is to learn a \emph{surrogate model} of the objective function~\cite{bartz2017model, ong2003evolutionary}, that is a data-driven approximation of the objective function that can be used in lieu of the objective function.

The most popular theoretical framework for surrogate-based optimization is currently ``Bayesian optimization''~\cite{shahriari2015taking,brochu2010tutorial}. Most instantiations model the objective function with Gaussian processes~\cite{williams2006gaussian} because (1) they are designed to give uncertainty estimates and (2) their explicit smoothness assumption allows them to make accurate predictions when little data is available. Once the model is defined, an \emph{acquisition function} is used to select the next solution to evaluate on the expensive objective function; this function typically uses the uncertainty estimates to balance exploration --- trying candidates in uncertain regions --- and exploitation --- trying candidates that are in the most promising regions according to the approximate model. In essence, Bayesian optimization loops over three steps: (1) finding the optimum of the acquisition function, which is a non-convex optimization problem with a ``cheap'' objective function, (2) evaluating the selected solution on the expensive function, and (3) updating the model with the new point. The model is usually initialized with a few candidates that are randomly chosen and evaluated on the expensive function.

The concepts of Bayesian optimization are easy to transfer to QD algorithm: the same models and the same model/optimization loop can be used. The only difference lies in the acquisition function (and how it is optimized), since the algorithm is not trying to find the optimum of a function anymore.

In the first experiments with surrogate-based QD algorithms, Gaier et al.~\cite{gaier2018data,gaier2017data,gaier2017aerodynamic} took inspiration from the Upper Confidence Bound (UCB)~\cite{cox1992statistical,srinivas2010gaussian}, which is a simple but successful acquisition function in Bayesian optimization~\cite{shahriari2015taking,brochu2010tutorial}:
\begin{equation}
    \textrm{UCB}(\boldsymbol{\theta}) = \mu(\boldsymbol{\theta}) + \beta \sigma(\boldsymbol{\theta})
\end{equation}
where $\mu(\boldsymbol{\theta})$ is the mean prediction according to the Gaussian processes (the surrogate model), $\sigma(\boldsymbol{\theta})$ is the prediction of the variance (which represents the uncertainty here). Intuitively, the optima of the UCB function are regions that are predicted as having a high predicted value ($\mu(\boldsymbol{\theta})$) and a high uncertainty ($\sigma(\boldsymbol{\theta})$). $\beta$ tunes the exploration-exploitation trade-offs (a large $\beta$ will favor candidates that have a high uncertainty, a small $\beta$ the candidates that have the highest predictions).

In Bayesian optimization, the algorithm would optimize the UCB according to the model and evaluate the optimum solution on the true objective function. However, in QD algorithms there is no single, most promising solution. Instead, Gaier et al. used MAP-Elites with the UCB as the objective function. This gives a ``acquisition map'' instead of a maximum of an acquisition function, that is, MAP-Elites outputs the candidate with best UCB value for each bin. In the spirit of MAP-Elites, and because no bin is considered more important than the others, Gaier et al. select candidates to be evaluated on the expensive function uniformly from this ``acquisition map''.

The resulting algorithm, called Surrogate-Assisted Illumination (SAIL), has been evaluated on the optimization of airfoils (minimizing drags for a given lift)~\cite{gaier2018data,gaier2017data,gaier2017aerodynamic}. The results show that with the same number of evaluations required by CMA-ES to find a near-optimal solution in a single bin (without a surrogate), SAIL finds solutions of similar quality in every bin (625 bins in theses experiments); in addition, when CMA-ES is used with a new surrogate for each bin, it requires an order of magnitude more evaluations than SAIL. Gaier et al. also showed promising results in a more complex 3-dimensional aerodynamic optimization in~\cite{gaier2018data}. In these experiments, the authors assumed that the bin was known from the values of a candidate solution (which is true in the design problem that they explored), but they suggest that the mean prediction of a second GP can be used to compute the coordinate sof the bin when creating the acquisition map (the variance of this prediction would be ignored).

In recent work, Kent and Branke proposed a different acquisition function that captures in a single value the expected improvement to the whole map, called the Expected Joint Improvement of Elites (EJIE)~\cite{kent2020bop}. Their starting point is the Expected Improvement (EI) acquisition function, which is another popular acquisition function in Bayesian optimization~\cite{jones1998efficient}. However, the expected improvement is defined relative to the best known value, which does not make sense in a QD algorithm. Instead, Kent and Branke propose to compute the expected improvement for each niche, then sum all the values, so that the acquisition function is maximum for points that are likely to provide large improvements to the objective value of one or more niches. In addition, they note that it is usually not possible to know the cell from the values of the candidate solution (an hypothesis that was made in the experiments with the SAIL algorithm): a second Gaussian process is needed to predict the bin for each candidate. This second process is integrated in the Expected Joint Improvement of Elites (EJIE) by weighting the expected improvement by the probability for a candidate to be in a given cell:
\begin{equation}
    \textrm{EJIE}(\boldsymbol{\theta}) = \sum_{i=1}^C P(\boldsymbol{\theta} \in c_i) EI_{c_i}(\boldsymbol{\theta})
\end{equation}
where $i$ is the bin index, $P(\boldsymbol{\theta} \in c_i)$ is the probability of $\boldsymbol{\theta}$ to be in bin $c_i$ (computed with a Gaussian process that models the features) and $EI_{c_i}(\boldsymbol{\theta})$ is the expected improvement for the bin $c_i$ (computed with a Gaussian process that models the objective function). For now, the EJIE acquisition gave promising results for a 1-dimensional benchmark problem~\cite{kent2020bop} and more work is needed to compare it to SAIL.

An interesting side-effect of modeling the objective function with a surrogate model is that it becomes easy to increase the resolution of the map by running a QD algorithm using only the models, which can usually be achieved in minutes.
\subsection{High-Dimensional Feature Space}
\label{sec:high_dim_qd}
The grid-based container approach of MAP-Elites has various benefits, such as conceptual simplicity, easy implementation, and it can form the basis for both quantitative QD metrics (e.g., number of filled bins, or the QD score~\cite{pugh2015confronting}), as well as qualitative evaluation (by visual inspection of a 2D map). For creating the grid, the user needs to provide a number of discretization intervals per feature dimension. This, however, has the drawback of not scaling to high-dimensional feature spaces, as the number of bins increases exponentially with the feature dimensions. For instance, for a 50-dimensional space and 2 discretizations per dimension, MAP-Elites would create an empty matrix of $1.13 \times 10^{15}$ cells that requires 4 petabytes of memory (assuming 4 bytes for the pointer of each cell). This motivates the question whether grid-based containers can be used in high-dimensional feature spaces.

\begin{figure}
    \centering
    \includegraphics[width=0.7\linewidth]{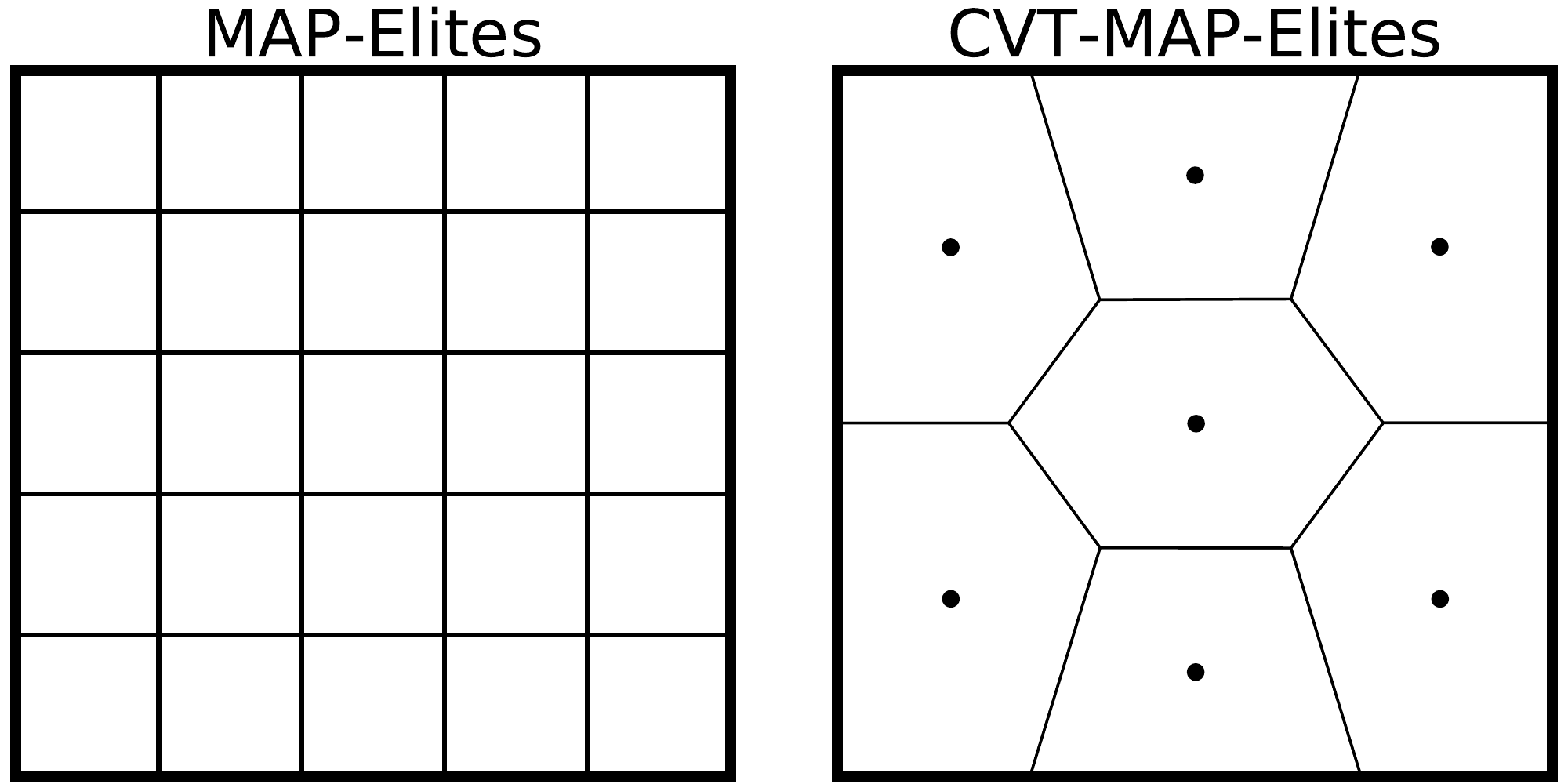}
    \caption{MAP-Elites (left) partitions the feature space using a number of bins (25 in this example), $k$, that is computed by the number of discretization intervals per dimension requested by the user (i.e., $k=\prod_{i=1}^{d}n_i$, where $d$ is the dimensionality of the feature space and $n_i$ is the discretization interval of dimension $i$). CVT-MAP-Elites (right) has explicit control over the number of bins (7 in this example) irrespective of the dimensionality of the feature space. Figure taken from~\cite{vassiliades2018using}. }\label{fig_cvt}
\end{figure}

Vassiliades et al.~\cite{vassiliades2018using} proposed an extension of MAP-Elites that addresses its dimensionality limitation using a tool from computational geometry known as a Centroidal Voronoi Tessellation (CVT)~\cite{du1999centroidal}. The key insight is that MAP-Elites defines the feature space using a bounding box that contains well-spread rectangular regions (Fig.~\ref{fig_cvt} left). A similar partitioning can be achieved using a CVT with the important difference that the number of regions is explicitly controlled, while the resulting regions have a convex polygonal shape (Fig.~\ref{fig_cvt} right).

The resulting CVT-MAP-Elites algorithm~\cite{vassiliades2018using}, requires a first, offline step (before QD optimization) to compute an \emph{approximate} CVT based on a user-provided number $k$ which defines the capacity of the container. The approximate CVT computation typically involves creating a dataset of $K >> k$ uniformly-distributed random points in the bounded feature space, and using the $k$-means clustering algorithm~\cite{macqueen1967kmeans} on this dataset to find $k$ centroids~\cite{ju2002probabilistic} (Algo.~\ref{algo:cvt}). If $K$ is large enough, the $k$ centroids become well-spread in the bounding volume. Typically as the number of dimensions increases so should $K$, however, the approximate nature of the algorithm, as well as the large number of solutions requested ($k$) by QD algorithms makes the algorithm perform well. For instance, for finding 10,000 effective gaits for a 6-legged robot, Vassiliades et al.~\cite{vassiliades2018using} used as feature space, subsets of the 36 parameters that define the gait~\cite{cully2015robots} (i.e., 12, 24 and 36), and demonstrated that CVT-MAP-Elites has the same performance irrespective of the dimensionality, in contrast to MAP-Elites.
\begin{algorithm}
    \caption{CVT approximation}\label{algo:cvt}
  \begin{algorithmic}[1]
    \Procedure{CVT}{$k, K$}
    \State $D \longleftarrow $ sample\_points($K$) \Comment{$K$ random samples}
    \State $\mathcal{C} \longleftarrow$ kmeans(D, k) \Comment{cluster dataset D using $k$ centroids}
    \State \textbf{return} centroids $\mathcal{C}$
    \EndProcedure
  \end{algorithmic}
\end{algorithm}

It is important to note that special care needs to be taken when the high-dimensional feature space is defined by sequences. Vassiliades et al.~\cite{vassiliades2018using} report experiments with a simulated maze-navigating mobile robot, and feature spaces of up-to 1000 dimensions composed by trajectories of the robot's $(x,y)$ location. In order for CVT-MAP-Elites to work, it needs representative centroids, and sampling from a uniform distribution (Algo.~\ref{algo:cvt} line~2) would not work, as it assumes that the various dimensions are independent. For these experiments, the authors~\cite{vassiliades2018using} used knowledge about the robot's physical constraints (i.e., it can move at most 2 units and it cannot exceed the bounds of the maze) in order to effectively sample random trajectories that cover the space well.

Another related issue (that can also apply to lower dimensional feature spaces) is when the bounds of the various feature space dimensions are not known \emph{a priori}. If wrong values are used, the performance of (CVT-)MAP-Elites might deteriorate as it will not be able to fill the grid (the size of the bins would either be too small or too big). A natural way of dealing with this issue is to let the feature descriptors of the sampled solutions define these bounds. Variants of MAP-Elites and CVT-MAP-Elites have explored this direction~\cite{vassiliades2017comparison} and allow for the expansion of the bounding box that defines the feature space.

Archive-based containers can naturally be used when the feature space is high-dimensional, as they calculate distances between the sampled solutions. However, algorithms that use such containers, for example, NSLC~\cite{lehman2011evolving}, often come with various parameters that need to be tweaked for the task at hand. An approach called Cluster-Elites~\cite{vassiliades2017comparison}, that blends ideas from CVT-MAP-Elites and archive-based containers, allows the generation of centroids in non-convex spaces based on the sampled solutions.
\subsection{Learning the Behavior Descriptor}
\label{sec:autonomous_bd}
%
Defining the behavioral descriptor space remains a challenging and
crucial aspect in QD algorithms, as this decision
determines the shape of the produced collections. It often requires a
certain level of expertise or prior knowledge on the task at hand to
know what are the features that the solutions can exhibit to
define the corresponding behavioral descriptor space. However, in
certain situations, this prior knowledge is not available or a user
might target specific subsets of the possible solutions according to
some conditions that are not easy to define in practice.  Several
pieces of work have proposed solutions to make easier and more automatic
the definition of the behavioral descriptor using learning
algorithms.
\begin{figure*}
    \centering
    \includegraphics[width=0.7\textwidth]{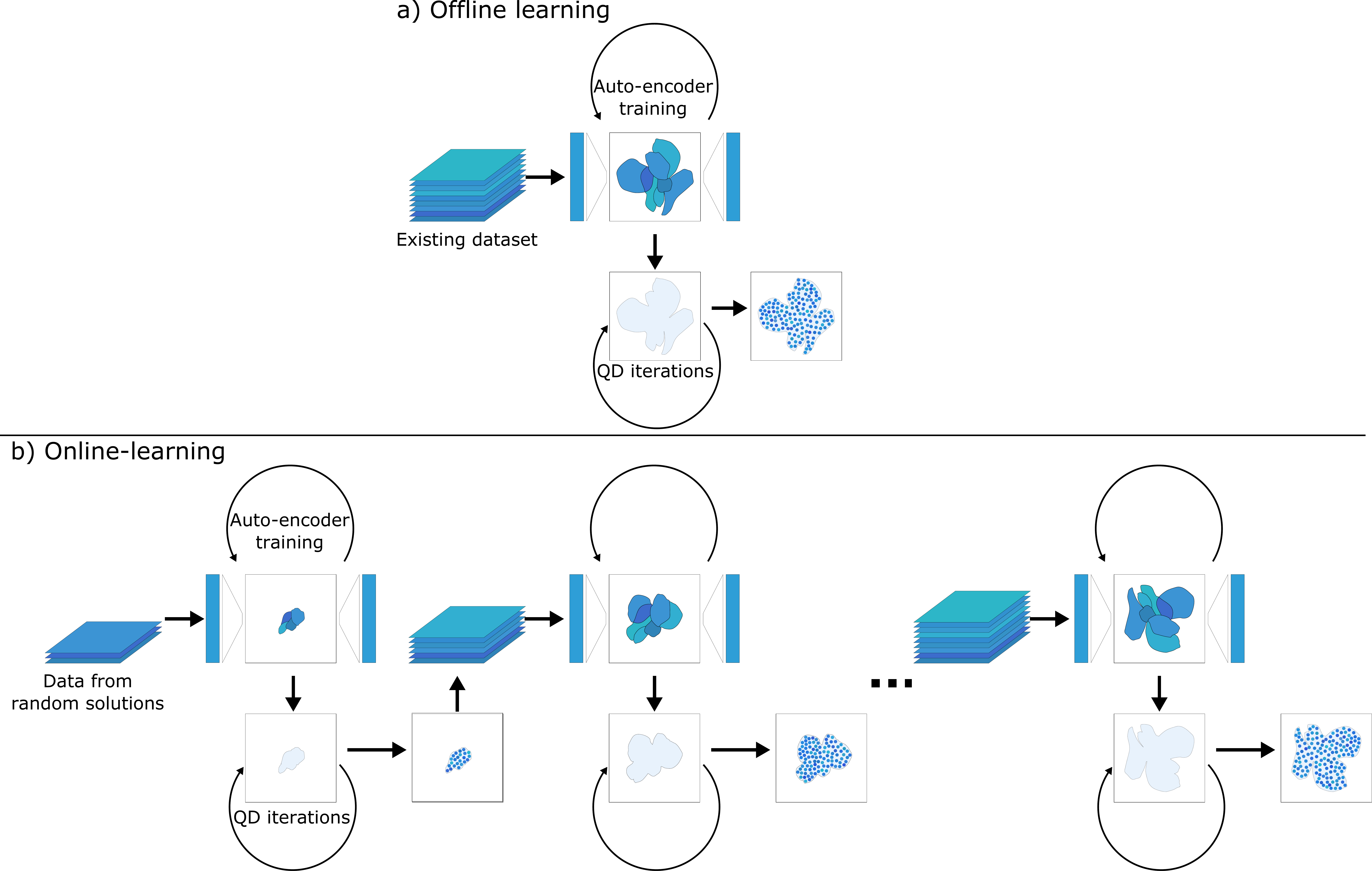}
    \caption{Different approaches to learning the behavioral descriptor.
      (\textbf{a}) An existing dataset can be used to train an
      auto-encoder (offline) and define a latent representation of the
      data. This latent space is then used as the behavioral
      descriptor space and a QD algorithm can build a collection of
      solutions filling this latent space. (\textbf{b}) Alternatively,
      the learning of the behavioral descriptor can be achieved
      during the QD optimization process. Starting from randomly
      generated solutions, the auto-encoder is trained and the
      produced latent space used a behavioral descriptor. Then, the
      several QD steps are executed to generates new solutions in the
      latent space, which increase the amount of data available to
      train the auto-encoder. The training of the auto-encoder is then
      extended and the process can repeat until convergence of both
      the QD algorithm and the auto-encoder.}\label{fig_learning_bd}
\end{figure*}

%
Sometimes the range of possible or desired types of solutions is known in
advance, but there is no easy way to programmatically encode this
knowledge into a low-dimensional BD definition. This is, for instance, the case when one wishes to generate
solutions resembling a set of examples that has been obtained from
a different process. A solution to address this problem is to collect all the possible features of each
example into a dataset which is then used to train a
dimensionality reduction algorithm, such as an auto-encoder or
principal component analysis. The outcome of this training is a
low-dimensional latent space that captures the relations and
similarities of the different examples and that can serve as a
behavioral descriptor space. The encoder in the case of the
auto-encoder (or the projector for the PCA) algorithm can project any
new solution in the descriptor space. With this learned behavioral
descriptor space, a QD algorithm produces a collection of solution that
covers the latent space and that maximize a fitness function. The
fitness function can be totally unrelated to the dimensionality
reduction algorithm, or it can be defined as the reconstruction error
of the algorithm, to encourage the QD algorithm to generate solutions
that look similar to the example set. A schematic illustration of this
approach is provided in Figure~\ref{fig_learning_bd}-a.

For example, this approach has been used to teach a robot to execute
trajectories that resemble handwritten digits. Finding a way to
characterise an arbitrary trajectory into a low-dimensional
behavioral descriptor capturing the diversity of hand-written digit
is particularly challenging. To side-step this challenge, Cully and
Demiris~\cite{cully2018hierarchical} employed an existing dataset of handwritten digits (the
well-known MNIST dataset~\cite{lecun1998gradient}) and train an auto-encoder to find a
low-dimensional latent space that captures the most prominent features of
the different digits. This latent space can then be used as the
behavioral descriptor space and QD algorithms produce
a collection of solutions that covers the space of the learned
features. As a result, this enabled a robotic arm to learn how to draw
digits without having to manually define what are the main features of
the different digits.

%
Instead of using an existing dataset, an alternative is to directly use
the solutions generated by the QD algorithm. In this case, the
randomly generated samples used in the initialisation of the QD algorithm
are used to train the dimensionality reduction algorithm. The
corresponding latent space definition becomes the behavioral
descriptor space for the execution of a few steps of the QD process.
during these steps, the number of solutions in the collection grows,
which has the direct effect of accumulating new samples that can be
used to extend the training of the dimensionality reduction algorithm
and thus change the definition of the latent space. The latent space
is then redefined to better include the new solutions that have been
discovered and redistribute them in the latent space according to
their high-level features. This process can be repeated periodically
to enable both the QD algorithm and the auto-encoder to
converge simultaneously. A schematic illustration of this approach is
provided in Figure~\ref{fig_learning_bd}-b.

The AURORA algorithm (AUtonomous RObots that Realise their Abilities,
\cite{cully2019autonomous}) uses this approach to enable robots to
discover large collections of diverse skills without any prior
knowledge on the capabilities of the robots. After each solution
evaluation, the trajectories of objects in the environment are
recorded and collected for the training of the auto-encoder. This
resulted in a collection of solutions to interact differently with
objects in the robot's environment. This concept has been extended in
the TAXONS~\cite{paolo2019unsupervised} algorithm by using raw images
from cameras placed on top of the robot and to discover how to move or
interact in the environment. A similar concept has also been used in
the DeLeNoX algorithm~\cite{liapis2013transforming} to evolve a
diversity of spacecraft sprite for video games using the
Novelty-Search algorithm.
\subsection{Improving Variation Operators}
\begin{figure}
    \includegraphics[width=\linewidth]{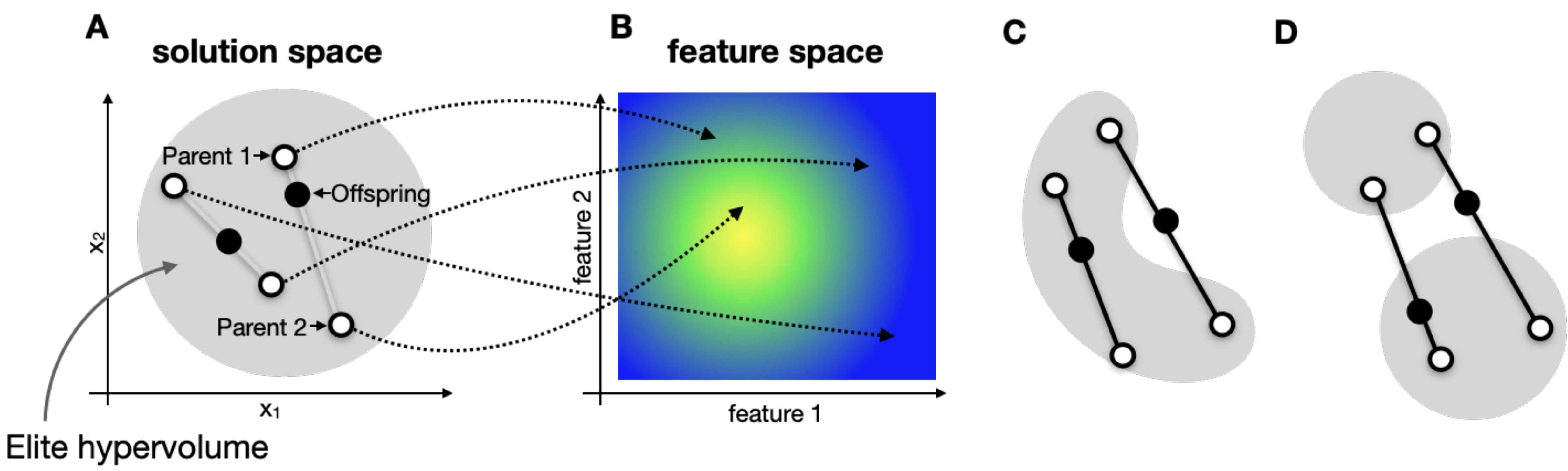}
    \caption{\label{fig:hypervolume}\textbf{Concept of the elite hypervolume~\cite{vassiliades2018discovering}}. \textbf{(A-B)} The elites of the feature space (B) occupy a very specific volume in the search space (A), here represented with a circle. If this volume is convex, then blending two elites creates a new elite. \textbf{(C-D)} If the elite hypervolume is not convex, then the blending is not guaranteed to generate a new elite, but it can still happen.}
\end{figure}
\label{sec:high_dim_genotype}
The state-of-the-art black-box optimization algorithms exploit the distribution of the highest-performing candidate solutions~\cite{hauschild2011introduction}; most notably, CMA-ES~\cite{hansen2001completely} computes a covariance matrix to sample the next population in the most promising direction. By contrast, the first QD algorithms used simple Gaussian variation (mutation) to generate new solutions, because they were focused on the selective pressure. However, the success of modern black-box optimization algorithms suggest that there is much to be improved by exploiting the distribution of the best candidate solutions in the variation operators.

Vassiliades and Mouret investigated how the genotypes of the archives of MAP-Elites are distributed in several benchmark tasks~\cite{vassiliades2018discovering}: while the elites are well spread in the feature space (by construction), they occupy a specific volume in the genotype space, that they called the ``Elite Hypervolume''. This should not be surprising because high-performing solutions often share ``similar recipes''. This echoes the high number of genes that are shared by species that live in very different niches, that is, that are elites of their ecological niche: for example, fruit flies and humans share about 60 percent of their genes~\cite{adams2000genome} (which correspond to neurons, muscle cells, etc.).

A straightforward way to exploit this hypervolume is to use the classic cross-over operator from evolutionary computation. The general idea is that a good variation operator is an operator that is likely to produce a high-performing solution from one or several existing high-performing solutions. In a QD algorithm, this means that we want to select one or several elites, which are in the current approximation of the elite hyper-volume, and create a solution in the elite hypervolume. If the hypervolume is convex, then any weighted average of elites will be in the hypervolume (Fig. \ref{fig:hypervolume}); if the volume is not convex, it might be locally convex and a ``blend'' of elites is still more likely to be in the elite hypervolume than not.

As a consequence, cross-over is surprisingly effective in QD algorithm, whereas its utility is more controversial in black-box optimization. For instance, evolutionary strategies like CMA-ES~\cite{hansen2001completely} do not use it all. Classic cross-over operators like SBX often work well~\cite{deb2001self}, but Vassiliades and Mouret proposed a simplified cross-over that gives good results in many QD problems, called ``directional variation''. Given two random elites $\boldsymbol{\theta}_{i}^{(t)}$ and $\boldsymbol{\theta}_{j}^{(t)}$, a new candidate solution $\boldsymbol{\theta}_{i}^{(t+1)}$ is generated by:
\begin{equation}
    \label{eq:line}
  \boldsymbol{\theta}_{i}^{(t+1)}=\boldsymbol{\theta}_{i}^{(t)}+\sigma_{1} \mathcal{N}(0, \mathbf{I})+\sigma_{2}\left(\boldsymbol{\theta}_{j}^{(t)}-\boldsymbol{\theta}_{i}^{(t)}\right) \mathcal{N}(0,1)
\end{equation}
where $\sigma_{1}$ controls the variance of the isotropic Gaussian distribution and $\sigma_{2}$ controls the magnitude of the perturbation of $\boldsymbol{\theta}_{i}^{(t)}$ along the direction of correlation with $\boldsymbol{\theta}_{j}^{(t)}$. 

During the same time, an independent study by Nordmoen et al.~\cite{nordmoen2018dynamic} investigated how static and dynamic mutation rates affect the exploration vs exploitation ability of MAP-Elites in a quadruped locomotion task. They tested three dynamic mutation schemes, the isotropic self-adaptation scheme of evolution strategies~\cite{beyer2002evolution}, one based on simulated annealing~\cite{kirkpatrick1983optimization} and another based on the QD coverage metric and found them to have either similar or better QD performance than the static mutation rates. Interestingly, the study by Vassiliades and Mouret~\cite{vassiliades2018discovering} also compared isotropic self-adaptation and found it to be less effective than their directional variation operator.

While crossover is a straightforward way to generate individuals that are likely to be in the elite hypervolume, it might sometimes be possible to fit a distribution so that we can directly sample from it. Since this volume can have any shape, modeling it with a simple Gaussian distribution is unlikely to be successful; instead, Gaier et al.~\cite{gaier2020discovering} proposed to use a Variational Auto-Encoder (VAE)~\cite{vae} that is learned from the genotype of the current archive of a QD algorithm. By learning the ``latent space'' of the data, the algorithm learns the non-linear ``recipes'' that define the high-performing solutions, so that more elites can be produced.

However, if the algorithm uses a Variational Auto-Encoder as a variation operator, it cannot generate candidates that are not in the current approximation of the hypervolume. In other words, when the QD algorithm is running, it can apply the current ``current recipe'' but not discover a ``better recipe'' for elites. It is therefore important to balance exploration -- trying candidates outside of the current distribution --- and exploitation --- learning and using the current recipe. To tackle this issue, Gaier et al. implemented a multi-armed bandit algorithm~\cite{auer2002finite,karafotias2014parameter} that tunes the probability of using the VAE and the probability of using other variation operators like Gaussian mutation and directional variation. As a result, the VAE is used only when it helps.

Gaier et al. tested this approach, called Data-Driven Encoding (DDE), in tasks up to 1000-dimensional~\cite{gaier2020discovering}. The results show that the VAE allows MAP-Elites to be used these kind high-dimensional task, although more tasks need to be investigated in the future. Interestingly, the latent space that is learned by the VAE is a representation of the elite hypervolume that can be reused for future optimization of tasks from the same distribution. For instance, it can be leveraged to quickly recompute a higher-dimensional map by using the latent space as the genotype space, or to fine-tune for a specific bin using a black-box optimizer like CMA-ES.

The combination of the uniform selection operator with standard mutation and crossover operators, which is used for instance in MAP-Elites, is a simple, yet very effective mechanism to produce new solutions. One of its strengths is that it is not biased toward any specific aspect of the optimization process. However, as a direct consequence, the exploration of the search space might be slower than alternative approaches. In particular, we can note that the selective pressure (i.e, the probability of a solution to be selected) induced by this mechanism is inversely proportional to the number of solutions in the collection, which can make this approach less appropriate for the generation of large collections.

Instead of such an unbiased selection mechanism, Fontaine et al.~\cite{fontaine2019covariance} proposed in the Covariance Matrix Adaptation MAP-Elites (CMA-ME) algorithm, the concept of \textit{emitters} that are independent processes in charge of the generation of new solutions. They defined three types of emitters, each based on the CMA-ES algorithm~\cite{hansen2001completely} to drive the exploration according to a specific intrinsic motivation. More precisely, they introduced the \textit{Optimizing emitters} that use CMA-ES to sample solutions that with higher performance, the \textit{Random Direction emitters} that favor solutions that are as far as possible along a randomly generated direction in the BD space, and the \textit{Improvement emitters} that reward solutions that are added to the collection (similarly to the curiosity score detailed above). These different emitters enable the user to specialise the search process for instance to accelerate the coverage of the behavioural space, or searching the nearest local optimum.

%
\subsection{Noisy Functions}
\label{sec:noisy}
One of the main current challenges in QD algorithm is noisy domains, in which the evaluation of the fitness and the behavioural descriptor values are subject to noise. This noise mainly perturbs the addition of solutions in the collection: a noisy BD measure might place a solution in a wrong cell (or region of the BD space), while a noisy fitness value might over-estimate the quality of the solution and lead to its undeserved preservation in the collection. A simple approach to overcome this challenge is to replicate the evaluation of each solution multiple times (e.g., 100 times) to collect statistics, such as the median for the fitness and the geometric median for the BD, to ensure a robust optimization process. However, this severely deteriorates the data-efficiency of the algorithm. To mitigate this problem and offer a solution that is both robust the noise and more data-efficient, Justesen et al.~\cite{justesen2019map} proposed to use the concept of \textit{adaptive sampling}~\cite{cantu2004adaptive} to allocate the evaluation budget only on solutions that are promising while avoiding the re-evaluations of solutions that are unlikely to be competitive. The results show that this approach is particularly effective when only the fitness function is noisy, as the noise on the BD creates \textit{drifting elites} (solutions that drifts to different cells and leaving behind them an empty cell).
An alternative approach is proposed by Flageat et al.~\cite{flageat2020fast} with the Deep-Grid MAP-Elites (DG-MAP-Elites), in which the grid of MAP-Elites is extended to store multiple solutions per cell, up to a fixed number (e.g., 50). This sub-population of solutions in each cell can be seen as the depth of the grid. Additionally, the selection and addition mechanisms have been changed to select with a higher probability high-performing  solutions within each subpopulation, while any new solution is added to the collection by replacing a randomly selected existing one. This avoids maintaining illegitimate solutions and forces the subpopulations to contain solutions that are likely to produce offspring that land in the same cell, thus improving the robustness of the BD.
\section{Conclusion}
\label{sec:conclusion}
Quality-Diversity Optimization is a novel branch of stochastic optimization that deals with a special kind of objective functions that return not only a fitness value, but also a behavior descriptor (or a feature vector). The goal of this type of optimization is to collect solutions in a container (or archive) so that they cover as much as possible the behavioral space, and they are locally optimized. We presented a short review of the history of QD optimization, and the main current topics under consideration in the community focusing more on the ones that we believe are the most promising directions to follow. Finally, we presented throughout the chapter many successful applications of QD algorithms on numerous fields, and gave an overview of the current limitations.

In particular, QD algorithms are very effective at (a) optimizing very sparse or non-convex functions, (b) illuminating the search space according to user-defined or learned features, and (c) producing locally optimized repertoires of behaviors (e.g. robots walking in every direction). We hope that readers of this chapter will now have an additional tool in their optimization toolkit, which will allow them to solve new problems or visualize their search spaces.
%
%
%
%
%
%
\bibliographystyle{abbrv}
\bibliography{references}{}

\begin{thebibliography}{80}
\providecommand{\natexlab}[1]{#1}
\providecommand{\url}[1]{\texttt{#1}}
\expandafter\ifx\csname urlstyle\endcsname\relax
  \providecommand{\doi}[1]{doi: #1}\else
  \providecommand{\doi}{doi: \begingroup \urlstyle{rm}\Url}\fi

\bibitem[Adams et~al.(2000)Adams, Celniker, Holt, Evans, Gocayne, Amanatides,
  Scherer, Li, Hoskins, Galle, et~al.]{adams2000genome}
Mark~D Adams, Susan~E Celniker, Robert~A Holt, Cheryl~A Evans, Jeannine~D
  Gocayne, Peter~G Amanatides, Steven~E Scherer, Peter~W Li, Roger~A Hoskins,
  Richard~F Galle, et~al.
\newblock The genome sequence of drosophila melanogaster.
\newblock \emph{Science}, 287\penalty0 (5461):\penalty0 2185--2195, 2000.

\bibitem[Auer et~al.(2002)Auer, Cesa-Bianchi, and Fischer]{auer2002finite}
Peter Auer, Nicolo Cesa-Bianchi, and Paul Fischer.
\newblock Finite-time analysis of the multiarmed bandit problem.
\newblock Springer, 2002.

\bibitem[Barrera and Coello(2009)]{barrera2009review}
Julio Barrera and Carlos A~Coello Coello.
\newblock A review of particle swarm optimization methods used for multimodal
  optimization.
\newblock In \emph{Innovations in swarm intelligence}, pages 9--37. Springer,
  2009.

\bibitem[Bartz-Beielstein and Zaefferer(2017)]{bartz2017model}
Thomas Bartz-Beielstein and Martin Zaefferer.
\newblock Model-based methods for continuous and discrete global optimization.
\newblock \emph{Applied Soft Computing}, 55:\penalty0 154--167, 2017.

\bibitem[Beyer and Schwefel(2002)]{beyer2002evolution}
Hans-Georg Beyer and Hans-Paul Schwefel.
\newblock Evolution strategies--a comprehensive introduction.
\newblock \emph{Natural computing}, 1\penalty0 (1):\penalty0 3--52, 2002.

\bibitem[Bradner et~al.(2014)Bradner, Iorio, and Davis]{Bradner2014}
Erin Bradner, Francesco Iorio, and Mark Davis.
\newblock {Parameters tell the design story: Ideation and abstraction in design
  optimization}.
\newblock \emph{Simulation Series}, 2014.
\newblock ISSN 07359276.

\bibitem[Brochu et~al.(2010)Brochu, Cora, and De~Freitas]{brochu2010tutorial}
Eric Brochu, Vlad~M Cora, and Nando De~Freitas.
\newblock A tutorial on bayesian optimization of expensive cost functions, with
  application to active user modeling and hierarchical reinforcement learning.
\newblock \emph{arXiv preprint arXiv:1012.2599}, 2010.

\bibitem[Cant{\'u}-Paz(2004)]{cantu2004adaptive}
Erick Cant{\'u}-Paz.
\newblock Adaptive sampling for noisy problems.
\newblock In \emph{Genetic and Evolutionary Computation Conference}, pages
  947--958. Springer, 2004.

\bibitem[Cavicchio(1970)]{cavicchio1970adaptive}
Daniel~Joseph Cavicchio.
\newblock \emph{Adaptive search using simulated evolution}.
\newblock PhD thesis, University of Michigan, Ann Arbor, MI, 1970.

\bibitem[Chatzilygeroudis et~al.(2018)Chatzilygeroudis, Vassiliades, and
  Mouret]{chatzilygeroudis2018reset}
Konstantinos Chatzilygeroudis, Vassilis Vassiliades, and Jean-Baptiste Mouret.
\newblock Reset-free trial-and-error learning for robot damage recovery.
\newblock \emph{Robotics and Autonomous Systems}, 100:\penalty0 236--250, 2018.

\bibitem[Clune et~al.(2013)Clune, Mouret, and Lipson]{clune2013evolutionary}
Jeff Clune, Jean-Baptiste Mouret, and Hod Lipson.
\newblock The evolutionary origins of modularity.
\newblock \emph{Proceedings of the Royal Society b: Biological sciences},
  280\penalty0 (1755):\penalty0 20122863, 2013.

\bibitem[Cox and John(1992)]{cox1992statistical}
Dennis~D Cox and Susan John.
\newblock A statistical method for global optimization.
\newblock In \emph{International Conference on Systems, Man, and Cybernetics},
  pages 1241--1246. IEEE, 1992.

\bibitem[Cully(2019)]{cully2019autonomous}
Antoine Cully.
\newblock Autonomous skill discovery with quality-diversity and unsupervised
  descriptors.
\newblock In \emph{Proceedings of the Genetic and Evolutionary Computation
  Conference}, pages 81--89. ACM, 2019.
\newblock \doi{10.1145/3321707.3321804}.

\bibitem[Cully and Demiris(2018{\natexlab{a}})]{cully2018hierarchical}
Antoine Cully and Yiannis Demiris.
\newblock Hierarchical behavioral repertoires with unsupervised descriptors.
\newblock \emph{Proceedings of the Genetic and Evolutionary Computation
  Conference}, 2018{\natexlab{a}}.

\bibitem[Cully and Demiris(2018{\natexlab{b}})]{cully2018quality}
Antoine Cully and Yiannis Demiris.
\newblock Quality and diversity optimization: A unifying modular framework.
\newblock \emph{IEEE Transactions on Evolutionary Computation}, 22\penalty0
  (2):\penalty0 245--259, 2018{\natexlab{b}}.

\bibitem[Cully and Mouret(2013)]{cully2013behavioral}
Antoine Cully and Jean-Baptiste Mouret.
\newblock Behavioral repertoire learning in robotics.
\newblock In \emph{Proceedings of the 15th annual conference on Genetic and
  Evolutionary Computation}, pages 175--182. ACM, 2013.

\bibitem[Cully et~al.(2015)Cully, Clune, Tarapore, and Mouret]{cully2015robots}
Antoine Cully, Jeff Clune, Danesh Tarapore, and Jean-Baptiste Mouret.
\newblock Robots that can adapt like animals.
\newblock \emph{Nature}, 521\penalty0 (7553):\penalty0 503--507, 2015.

\bibitem[Das et~al.(2011)Das, Maity, Qu, and Suganthan]{das2011real}
Swagatam Das, Sayan Maity, Bo-Yang Qu, and Ponnuthurai~Nagaratnam Suganthan.
\newblock Real-parameter evolutionary multimodal optimization — a survey of
  the state-of-the-art.
\newblock \emph{Swarm and Evolutionary Computation}, 1:\penalty0 71--88, 2011.

\bibitem[De~Jong(1975)]{de1975analysis}
Kenneth~Alan De~Jong.
\newblock \emph{Analysis of the behavior of a class of genetic adaptive
  systems}.
\newblock PhD thesis, University of Michigan, Ann Arbor, MI, 1975.

\bibitem[Deb and Beyer(2001)]{deb2001self}
Kalyanmoy Deb and Hans-Georg Beyer.
\newblock Self-adaptive genetic algorithms with simulated binary crossover.
\newblock \emph{Evolutionary computation}, 9\penalty0 (2):\penalty0 197--221,
  2001.

\bibitem[Deb and Saha(2010)]{deb2010finding}
Kalyanmoy Deb and Amit Saha.
\newblock Finding multiple solutions for multimodal optimization problems using
  a multi-objective evolutionary approach.
\newblock In \emph{Proceedings of the 12th annual conference on genetic and
  evolutionary computation}, pages 447--454, 2010.

\bibitem[Du et~al.(1999)Du, Faber, and Gunzburger]{du1999centroidal}
Qiang Du, Vance Faber, and Max Gunzburger.
\newblock Centroidal {Voronoi} tessellations: applications and algorithms.
\newblock \emph{SIAM review}, 41:\penalty0 637--676, 1999.

\bibitem[Duarte et~al.(2018)Duarte, Gomes, Oliveira, and
  Christensen]{duarte2018evolution}
Miguel Duarte, Jorge Gomes, Sancho~Moura Oliveira, and Anders~Lyhne
  Christensen.
\newblock Evolution of repertoire-based control for robots with complex
  locomotor systems.
\newblock \emph{IEEE Transactions on Evolutionary Computation}, 22\penalty0
  (2):\penalty0 314--328, 2018.

\bibitem[Ecoffet et~al.(2019)Ecoffet, Huizinga, Lehman, Stanley, and
  Clune]{ecoffet2019go}
Adrien Ecoffet, Joost Huizinga, Joel Lehman, Kenneth~O Stanley, and Jeff Clune.
\newblock Go-explore: a new approach for hard-exploration problems.
\newblock \emph{arXiv preprint arXiv:1901.10995}, 2019.

\bibitem[Ecoffet et~al.(2020)Ecoffet, Huizinga, Lehman, Stanley, and
  Clune]{ecoffet2020first}
Adrien Ecoffet, Joost Huizinga, Joel Lehman, Kenneth~O Stanley, and Jeff Clune.
\newblock First return then explore.
\newblock \emph{arXiv preprint arXiv:2004.12919}, 2020.

\bibitem[Escande et~al.(2014)Escande, Mansard, and
  Wieber]{escande2014hierarchical}
Adrien Escande, Nicolas Mansard, and Pierre-Brice Wieber.
\newblock Hierarchical quadratic programming: Fast online humanoid-robot motion
  generation.
\newblock \emph{The International Journal of Robotics Research}, 33\penalty0
  (7):\penalty0 1006--1028, 2014.

\bibitem[Flageat and Cully(2020)]{flageat2020fast}
Manon Flageat and Antoine Cully.
\newblock Fast and stable map-elites in noisy domains using deep grids.
\newblock \emph{Proceeding of the Alife conference}, 2020.

\bibitem[Fontaine et~al.(2020)Fontaine, Togelius, Nikolaidis, and
  Hoover]{fontaine2019covariance}
Matthew~C Fontaine, Julian Togelius, Stefanos Nikolaidis, and Amy~K Hoover.
\newblock Covariance matrix adaptation for the rapid illumination of behavior
  space.
\newblock \emph{Proceedings of the Genetic and Evolutionary Computation
  Conference Companion}, 2020.

\bibitem[Gaier et~al.(2017{\natexlab{a}})Gaier, Asteroth, and
  Mouret]{gaier2017aerodynamic}
Adam Gaier, Alexander Asteroth, and Jean-Baptiste Mouret.
\newblock Aerodynamic design exploration through surrogate-assisted
  illumination.
\newblock In \emph{18th AIAA/ISSMO Multidisciplinary Analysis and Optimization
  Conference}, page 3330, 2017{\natexlab{a}}.

\bibitem[Gaier et~al.(2017{\natexlab{b}})Gaier, Asteroth, and
  Mouret]{gaier2017data}
Adam Gaier, Alexander Asteroth, and Jean-Baptiste Mouret.
\newblock Data-efficient exploration, optimization, and modeling of diverse
  designs through surrogate-assisted illumination.
\newblock In \emph{Proceedings of the Genetic and Evolutionary Computation
  Conference}, pages 99--106. ACM, 2017{\natexlab{b}}.

\bibitem[Gaier et~al.(2018)Gaier, Asteroth, and Mouret]{gaier2018data}
Adam Gaier, Alexander Asteroth, and Jean-Baptiste Mouret.
\newblock Data-efficient design exploration through surrogate-assisted
  illumination.
\newblock \emph{Evolutionary computation}, pages 1--30, 2018.

\bibitem[Gaier et~al.(2020)Gaier, Asteroth, and Mouret]{gaier2020discovering}
Adam Gaier, Alexander Asteroth, and Jean-Baptiste Mouret.
\newblock Discovering representations for black-box optimization.
\newblock In \emph{Proceedings of the Genetic and Evolutionary Computation
  Conference (GECCO)}, volume~11, 2020.

\bibitem[Goldberg et~al.(1987)Goldberg, Richardson,
  et~al.]{goldberg1987genetic}
David~E Goldberg, Jon Richardson, et~al.
\newblock Genetic algorithms with sharing for multimodal function optimization.
\newblock In \emph{Genetic algorithms and their applications: Proceedings of
  the Second International Conference on Genetic Algorithms}, pages 41--49.
  Hillsdale, NJ: Lawrence Erlbaum, 1987.

\bibitem[Hansen and Ostermeier(2001)]{hansen2001completely}
Nikolaus Hansen and Andreas Ostermeier.
\newblock Completely derandomized self-adaptation in evolution strategies.
\newblock \emph{Evolutionary computation}, 9\penalty0 (2):\penalty0 159--195,
  2001.

\bibitem[Harik(1995)]{harik1995rts}
Georges~R Harik.
\newblock Finding multimodal solutions using restricted tournament selection.
\newblock In \emph{Proceedings of the 6th International Conference on Genetic
  Algorithms}, pages 24--31, San Francisco, CA, 1995. Morgan Kaufmann.

\bibitem[Hauschild and Pelikan(2011)]{hauschild2011introduction}
Mark Hauschild and Martin Pelikan.
\newblock An introduction and survey of estimation of distribution algorithms.
\newblock \emph{Swarm and evolutionary computation}, 1\penalty0 (3):\penalty0
  111--128, 2011.

\bibitem[Jones et~al.(1998)Jones, Schonlau, and Welch]{jones1998efficient}
Donald~R Jones, Matthias Schonlau, and William~J Welch.
\newblock Efficient global optimization of expensive black-box functions.
\newblock \emph{Journal of Global optimization}, 13\penalty0 (4):\penalty0
  455--492, 1998.

\bibitem[Ju et~al.(2002)Ju, Du, and Gunzburger]{ju2002probabilistic}
Lili Ju, Qiang Du, and Max Gunzburger.
\newblock Probabilistic methods for centroidal {Voronoi} tessellations and
  their parallel implementations.
\newblock \emph{Parallel Computing}, 28\penalty0 (10):\penalty0 1477--1500,
  2002.

\bibitem[Justesen et~al.(2019)Justesen, Risi, and Mouret]{justesen2019map}
Niels Justesen, Sebastian Risi, and Jean-Baptiste Mouret.
\newblock Map-elites for noisy domains by adaptive sampling.
\newblock In \emph{Proceedings of the Genetic and Evolutionary Computation
  Conference Companion}, pages 121--122. ACM, 2019.

\bibitem[Karafotias et~al.(2014)Karafotias, Hoogendoorn, and
  Eiben]{karafotias2014parameter}
Giorgos Karafotias, Mark Hoogendoorn, and {\'A}goston~E Eiben.
\newblock Parameter control in evolutionary algorithms: Trends and challenges.
\newblock \emph{IEEE Transactions on Evolutionary Computation}, 19\penalty0
  (2):\penalty0 167--187, 2014.

\bibitem[Kent and Branke(2020)]{kent2020bop}
Paul Kent and Juergen Branke.
\newblock Bop-elites, a bayesian optimisation algorithm for quality-diversity
  search.
\newblock \emph{arXiv preprint arXiv:2005.04320}, 2020.

\bibitem[Kingma and Welling(2014)]{vae}
Diederik~P. Kingma and Max Welling.
\newblock Auto-encoding variational {Bayes}.
\newblock In Yoshua Bengio and Yann LeCun, editors, \emph{International
  Conference on Learning Representation ({ICLR})}, 2014.

\bibitem[Kirkpatrick et~al.(1983)Kirkpatrick, Gelatt, and
  Vecchi]{kirkpatrick1983optimization}
Scott Kirkpatrick, C~Daniel Gelatt, and Mario~P Vecchi.
\newblock Optimization by simulated annealing.
\newblock \emph{science}, 220\penalty0 (4598):\penalty0 671--680, 1983.

\bibitem[LeCun et~al.(1998)LeCun, Bottou, Bengio, and
  Haffner]{lecun1998gradient}
Yann LeCun, L{\'e}on Bottou, Yoshua Bengio, and Patrick Haffner.
\newblock Gradient-based learning applied to document recognition.
\newblock \emph{Proceedings of the IEEE}, 86\penalty0 (11):\penalty0
  2278--2324, 1998.

\bibitem[Lee et~al.(1999)Lee, Cho, and Jung]{lee1999niching}
Cheol-Gyun Lee, Dong-Hyeok Cho, and Hyun-Kyo Jung.
\newblock Niching genetic algorithm with restricted competition selection for
  multimodal function optimization.
\newblock \emph{IEEE Transactions on Magnetics}, 35\penalty0 (3):\penalty0
  1722--1725, 1999.

\bibitem[Lehman and Stanley(2011{\natexlab{a}})]{lehman2011abandoning}
Joel Lehman and Kenneth~O Stanley.
\newblock Abandoning objectives: Evolution through the search for novelty
  alone.
\newblock \emph{Evolutionary computation}, 19\penalty0 (2):\penalty0 189--223,
  2011{\natexlab{a}}.

\bibitem[Lehman and Stanley(2011{\natexlab{b}})]{lehman2011evolving}
Joel Lehman and Kenneth~O Stanley.
\newblock Evolving a diversity of virtual creatures through novelty search and
  local competition.
\newblock In \emph{Proceedings of the 13th annual conference on Genetic and
  evolutionary computation}, pages 211--218. ACM, 2011{\natexlab{b}}.

\bibitem[Lehman et~al.(2016)Lehman, Risi, and Clune]{lehman2016creative}
Joel Lehman, Sebastian Risi, and Jeff Clune.
\newblock Creative generation of 3d objects with deep learning and innovation
  engines.
\newblock In \emph{Proceedings of the 7th International Conference on
  Computational Creativity}, 2016.

\bibitem[Liapis et~al.(2013)Liapis, Mart{\i}nez, Togelius, and
  Yannakakis]{liapis2013transforming}
Antonios Liapis, H{\'e}ctor~P Mart{\i}nez, Julian Togelius, and Georgios~N
  Yannakakis.
\newblock Transforming exploratory creativity with delenox.
\newblock In \emph{Proceedings of the Fourth International Conference on
  Computational Creativity}, pages 56--63. AAAI Press, 2013.

\bibitem[MacQueen(1967)]{macqueen1967kmeans}
James MacQueen.
\newblock Some methods for classification and analysis of multivariate
  observations.
\newblock In \emph{Proc. 5th {Berkeley} Symp. on Math. Statist. and Prob.},
  volume~1, pages 281--297, Berkeley, CA, 1967. Univ. of Calif. Press.

\bibitem[Mahfoud(1995)]{mahfoud1995niching}
Samir Mahfoud.
\newblock \emph{Niching methods for genetic algorithms}.
\newblock PhD thesis, University of Illinois at Urbana-Champaign, Urbana, IL,
  1995.

\bibitem[Mayne et~al.(2000)Mayne, Rawlings, Rao, and
  Scokaert]{mayne2000constrained}
David~Q Mayne, James~B Rawlings, Christopher~V Rao, and Pierre~OM Scokaert.
\newblock Constrained model predictive control: Stability and optimality.
\newblock Elsevier, 2000.

\bibitem[Mouret and Doncieux(2012)]{mouret2012encouraging}
J-B Mouret and St{\'e}phane Doncieux.
\newblock Encouraging behavioral diversity in evolutionary robotics: An
  empirical study.
\newblock \emph{Evolutionary computation}, 20\penalty0 (1):\penalty0 91--133,
  2012.

\bibitem[Mouret and Clune(2015)]{mouret2015illuminating}
Jean-Baptiste Mouret and Jeff Clune.
\newblock Illuminating search spaces by mapping elites.
\newblock \emph{arXiv preprint arXiv:1504.04909}, 2015.

\bibitem[Mouret and Maguire(2020)]{mouret2020quality}
Jean-Baptiste Mouret and Glenn Maguire.
\newblock Quality diversity for multi-task optimization.
\newblock In \emph{Proceedings of the Genetic and Evolutionary Computation
  Conference}. ACM, 2020.

\bibitem[Nguyen et~al.(2015{\natexlab{a}})Nguyen, Yosinski, and
  Clune]{nguyen2015deep}
Anh Nguyen, Jason Yosinski, and Jeff Clune.
\newblock Deep neural networks are easily fooled: High confidence predictions
  for unrecognizable images.
\newblock In \emph{Proceedings of the IEEE conference on computer vision and
  pattern recognition}, pages 427--436, 2015{\natexlab{a}}.

\bibitem[Nguyen et~al.(2015{\natexlab{b}})Nguyen, Yosinski, and
  Clune]{nguyen2015innovation}
Anh~Mai Nguyen, Jason Yosinski, and Jeff Clune.
\newblock Innovation engines: Automated creativity and improved stochastic
  optimization via deep learning.
\newblock In \emph{Proceedings of the 2015 Annual Conference on Genetic and
  Evolutionary Computation}, pages 959--966. ACM, 2015{\natexlab{b}}.

\bibitem[Nordmoen et~al.(2018)Nordmoen, Samuelsen, Ellefsen, and
  Glette]{nordmoen2018dynamic}
Jorgen Nordmoen, Eivind Samuelsen, Kai~Olav Ellefsen, and Kyrre Glette.
\newblock Dynamic mutation in map-elites for robotic repertoire generation.
\newblock In \emph{Artificial Life Conference Proceedings}, pages 598--605. MIT
  Press, 2018.

\bibitem[Ong et~al.(2003)Ong, Nair, and Keane]{ong2003evolutionary}
Yew~S Ong, Prasanth~B Nair, and Andrew~J Keane.
\newblock Evolutionary optimization of computationally expensive problems via
  surrogate modeling.
\newblock \emph{AIAA journal}, 41\penalty0 (4):\penalty0 687--696, 2003.

\bibitem[Paolo et~al.(2019)Paolo, Laflaquiere, Coninx, and
  Doncieux]{paolo2019unsupervised}
Giuseppe Paolo, Alban Laflaquiere, Alexandre Coninx, and Stephane Doncieux.
\newblock Unsupervised learning and exploration of reachable outcome space.
\newblock \emph{algorithms}, 24:\penalty0 25, 2019.

\bibitem[Pearce and Branke(2018)]{pearce2018continuous}
Michael Pearce and Juergen Branke.
\newblock Continuous multi-task bayesian optimisation with correlation.
\newblock \emph{European Journal of Operational Research}, 270\penalty0
  (3):\penalty0 1074--1085, 2018.

\bibitem[P{\'e}trowski(1996)]{petrowski1996clearing}
Alain P{\'e}trowski.
\newblock A clearing procedure as a niching method for genetic algorithms.
\newblock In \emph{Proceedings of IEEE international conference on evolutionary
  computation}, pages 798--803. IEEE, 1996.

\bibitem[Preuss(2015)]{preuss2015multimodal}
Mike Preuss.
\newblock \emph{Multimodal optimization by means of evolutionary algorithms}.
\newblock Springer, 2015.

\bibitem[Preuss et~al.(2005)Preuss, Sch{\"o}nemann, and
  Emmerich]{preuss2005counteracting}
Mike Preuss, Lutz Sch{\"o}nemann, and Michael Emmerich.
\newblock Counteracting genetic drift and disruptive recombination in ($\mu
  \overset{+}{,} \lambda$)-{EA} on multimodal fitness landscapes.
\newblock In \emph{Proceedings of the 7th annual conference on Genetic and
  evolutionary computation}, pages 865--872, 2005.

\bibitem[Pugh et~al.(2015)Pugh, Soros, Szerlip, and
  Stanley]{pugh2015confronting}
Justin~K Pugh, LB~Soros, Paul~A Szerlip, and Kenneth~O Stanley.
\newblock Confronting the challenge of quality diversity.
\newblock In \emph{Proceedings of the 2015 on Genetic and Evolutionary
  Computation Conference}, pages 967--974. ACM, 2015.

\bibitem[Pugh et~al.(2016)Pugh, Soros, and Stanley]{pugh2016quality}
Justin~K Pugh, Lisa~B Soros, and Kenneth~O Stanley.
\newblock Quality diversity: A new frontier for evolutionary computation.
\newblock \emph{Frontiers in Robotics and AI}, 3:\penalty0 40, 2016.

\bibitem[Rudolph(2001)]{rudolph2001self}
G{\"u}nter Rudolph.
\newblock Self-adaptive mutations may lead to premature convergence.
\newblock \emph{IEEE Transactions on Evolutionary Computation}, 5\penalty0
  (4):\penalty0 410--414, 2001.

\bibitem[Sareni and Krahenbuhl(1998)]{sareni1998fitness}
Bruno Sareni and Laurent Krahenbuhl.
\newblock Fitness sharing and niching methods revisited.
\newblock \emph{IEEE Transations on Evolutionary Computation}, 2:\penalty0
  97--106, 1998.

\bibitem[Shahriari et~al.(2015)Shahriari, Swersky, Wang, Adams, and
  De~Freitas]{shahriari2015taking}
Bobak Shahriari, Kevin Swersky, Ziyu Wang, Ryan~P Adams, and Nando De~Freitas.
\newblock Taking the human out of the loop: A review of bayesian optimization.
\newblock \emph{Proceedings of the IEEE}, 104\penalty0 (1):\penalty0 148--175,
  2015.

\bibitem[Shir et~al.(2007)Shir, Emmerich, B{\"a}ck, and
  Vrakking]{shir2007conceptual}
OM~Shir, M~Emmerich, Th~B{\"a}ck, and MJJ Vrakking.
\newblock Conceptual designs in laser pulse shaping obtained by niching in
  evolution strategies.
\newblock In \emph{EUROGEN 2007}, 2007.

\bibitem[Sigmund(2001)]{sigmund200199}
Ole Sigmund.
\newblock A 99 line topology optimization code written in matlab.
\newblock \emph{Structural and multidisciplinary optimization}, 21\penalty0
  (2):\penalty0 120--127, 2001.

\bibitem[Singh and Deb(2006)]{singh2006comparison}
Gulshan Singh and Kalyanmoy Deb.
\newblock Comparison of multi-modal optimization algorithms based on
  evolutionary algorithms.
\newblock In \emph{Proceedings of the 8th annual conference on Genetic and
  evolutionary computation}, pages 1305--1312, 2006.

\bibitem[Srinivas et~al.(2010)Srinivas, Krause, Kakade, and
  Seeger]{srinivas2010gaussian}
Niranjan Srinivas, Andreas Krause, Sham Kakade, and Matthias Seeger.
\newblock Gaussian process optimization in the bandit setting: no regret and
  experimental design.
\newblock In \emph{Proceedings of the 27th International Conference on
  International Conference on Machine Learning}, pages 1015--1022, 2010.

\bibitem[Tarapore et~al.(2016)Tarapore, Clune, Cully, and
  Mouret]{tarapore2016different}
Danesh Tarapore, Jeff Clune, Antoine Cully, and Jean-Baptiste Mouret.
\newblock How do different encodings influence the performance of the
  map-elites algorithm?
\newblock In \emph{Genetic and Evolutionary Computation Conference}, 2016.

\bibitem[Vassiliades and Mouret(2018)]{vassiliades2018discovering}
Vassilis Vassiliades and Jean-Baptiste Mouret.
\newblock Discovering the elite hypervolume by leveraging interspecies
  correlation.
\newblock \emph{Proceedings of the Genetic and Evolutionary Computation
  Conference}, 2018.

\bibitem[Vassiliades et~al.(2017{\natexlab{a}})Vassiliades, Chatzilygeroudis,
  and Mouret]{vassiliades2017comparing}
Vassilis Vassiliades, Konstantinos Chatzilygeroudis, and Jean-Baptiste Mouret.
\newblock Comparing multimodal optimization and illumination.
\newblock In \emph{Proceedings of the Genetic and Evolutionary Computation
  Conference Companion}, pages 97--98. ACM, 2017{\natexlab{a}}.

\bibitem[Vassiliades et~al.(2017{\natexlab{b}})Vassiliades, Chatzilygeroudis,
  and Mouret]{vassiliades2017comparison}
Vassilis Vassiliades, Konstantinos Chatzilygeroudis, and Jean-Baptiste Mouret.
\newblock A comparison of illumination algorithms in unbounded spaces.
\newblock In \emph{Proceedings of the Genetic and Evolutionary Computation
  Conference Companion}, pages 1578--1581. ACM, 2017{\natexlab{b}}.

\bibitem[Vassiliades et~al.(2018)Vassiliades, Chatzilygeroudis, and
  Mouret]{vassiliades2018using}
Vassilis Vassiliades, Konstantinos Chatzilygeroudis, and Jean-Baptiste Mouret.
\newblock Using centroidal voronoi tessellations to scale up the
  multidimensional archive of phenotypic elites algorithm.
\newblock \emph{IEEE Transactions on Evolutionary Computation}, 22\penalty0
  (4):\penalty0 623--630, 2018.

\bibitem[Williams and Rasmussen(2006)]{williams2006gaussian}
Christopher~KI Williams and Carl~Edward Rasmussen.
\newblock \emph{Gaussian processes for machine learning}, volume~2.
\newblock MIT press Cambridge, MA, 2006.

\bibitem[Yin and Germay(1993)]{yin1993fast}
Xiaodong Yin and Noel Germay.
\newblock A fast genetic algorithm with sharing scheme using cluster analysis
  methods in multimodal function optimization.
\newblock In \emph{Artificial neural nets and genetic algorithms}, pages
  450--457. Springer, 1993.

\end{thebibliography}
\end{document}